\documentclass[preprint,12pt]{elsarticle}

\usepackage{amssymb}
\usepackage{amsmath}

\usepackage[ruled]{algorithm2e}
\usepackage{array} 
\usepackage{multirow} 
\usepackage{wrapfig}
\usepackage{subcaption}
\usepackage{threeparttable}  
\usepackage[table]{xcolor} 
\usepackage{colortbl}
\usepackage[utf8]{inputenc} 
\usepackage[T1]{fontenc}    
\usepackage{hyperref}       
\usepackage{url}            
\usepackage{booktabs}       
\usepackage{amsfonts}       
\usepackage{nicefrac}       
\usepackage{microtype}      
\usepackage{xcolor}         
\usepackage{color}
\usepackage{pifont}
\usepackage{graphicx}
\usepackage{dsfont}
\usepackage{enumitem}
\usepackage{float}

\definecolor{ggray}{rgb}{0.85, 0.85, 0.85}
\definecolor{ggreen}{rgb}{0.14117, 0.529411, 0.35294}

\newcommand{\xf}[1]{{\color{black} #1}}
\newcommand{\green}[1]{{\color{ggreen} #1}}
\newcommand{\gray}[1]{{\color{ggray} #1}}
\journal{Information Fusion}

\begin{document}

\begin{frontmatter}



\title{\large Cues3D: Unleashing the Power of Sole NeRF for Consistent and Unique Instances in Open-Vocabulary 3D Panoptic Segmentation}




\author[label1,label2]{Feng Xue\fnref{equalcontribution}}\ead{feng.xue@unitn.it}
\author[label1]{Wenzhuang Xu\fnref{equalcontribution}}\ead{xuwenzhuang@bupt.edu.cn}
\author[label1]{Guofeng Zhong}\ead{zgf@bupt.edu.cn}
\author[label1]{\\Anlong Ming\corref{mycorrespondingauthor}}\ead{mal@bupt.edu.cn}
\author[label2]{Nicu Sebe}\ead{niculae.sebe@unitn.it}

\affiliation[label1]{organization={Beijing University of Posts and Telecommunications},
            city={Beijing},
            postcode={100876},
            country={China}}

\affiliation[label2]{organization={University of Trento},
            city={Trento},
            postcode={38123},
            country={Italy}}

\fntext[equalcontribution]{Equal contribution.}
\cortext[mycorrespondingauthor]{Corresponding author.}

\begin{abstract}
Open-vocabulary 3D panoptic segmentation has recently emerged as a significant trend.
Top-performing methods currently integrate 2D segmentation with geometry-aware 3D primitives.
However, the advantage would be lost without high-fidelity 3D point clouds,
such as methods based on Neural Radiance Field (NeRF).
\xf{These methods are limited by the insufficient capacity to maintain consistency across partial observations}.
To address this,
recent works have utilized contrastive loss or cross-view association pre-processing for view consensus.
In contrast to them,
we present \textbf{Cues3D},
a compact approach that relies solely on NeRF instead of pre-associations.
The core idea is that NeRF's implicit 3D field \xf{inherently establishes a globally consistent geometry,
enabling effective object distinction without explicit cross-view supervision.}
We propose a three-phase training framework for NeRF,
\textit{initialization}-\textit{disambiguation}-\textit{refinement},
whereby the instance IDs are corrected using the initially-learned knowledge.
Additionally, an instance disambiguation method is proposed to match NeRF-rendered 3D masks and ensure \xf{globally unique} 3D instance identities.
With the aid of Cues3D,
we obtain highly consistent and unique 3D instance ID for each object across views with a balanced version of NeRF.
Our experiments are conducted on ScanNet v2, ScanNet200, ScanNet++, and Replica datasets for 3D instance, panoptic, and semantic segmentation tasks.
Cues3D outperforms other 2D image-based methods and competes with the latest 2D-3D merging based methods,
while even surpassing them when using additional 3D point clouds.
The code link could be found in the appendix and will be released on \href{https://github.com/mRobotit/Cues3D} {\textit{github}}.
\end{abstract}



\begin{keyword}
Open-Vocabulary 3D Panoptic Segmentation \sep Multi-view Instance Consistency \sep Global Uniqueness of Instances
\end{keyword}

\end{frontmatter}




\begin{figure}[ht]
\centering
\includegraphics[width=1\textwidth]{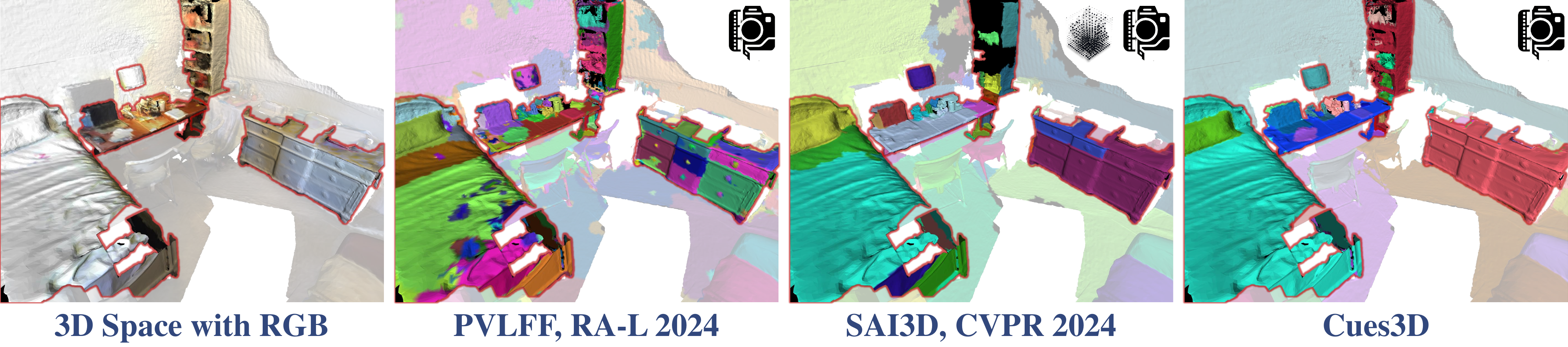}
\caption{\textbf{3D Visualization of the latest methods (aligned to the reconstructed point cloud).}
PVLFF \cite{pvlff} takes 2D images as input,
same as Cues3D.
SAI3D \cite{yin2024sai3d} uses both 2D/3D data.
Cues3D obtains better instance integrity around ``\textit{bed}'', ``\textit{cabinet}'', ``\textit{bookshelf}'', and ``\textit{table}'' area (highlighted).}
\label{fig:visualization_compare}
\end{figure}

\section{Introduction}
3D panoptic segmentation,
which aims to segment 3D masks of all instances (defined as ``\textbf{things}'') and background (defined as ``\textbf{stuffs}'') in scenes and recognize their categories,
is a key component of robotics and embodied AI \cite{WU2024102532,10160969} for building spatial semantic perception.
In the past year,
\textbf{open-vocabulary} perception solutions have gained increasing attention compared to traditional ones \cite{ICRA,liang2023unknown,xue_indoor_2023},
due to their ability to obtain semantic representations of objects that are not presented at all in the training set from vision-language models (VLMs).

\begin{figure}[t]
\centering
\includegraphics[width=1\textwidth]{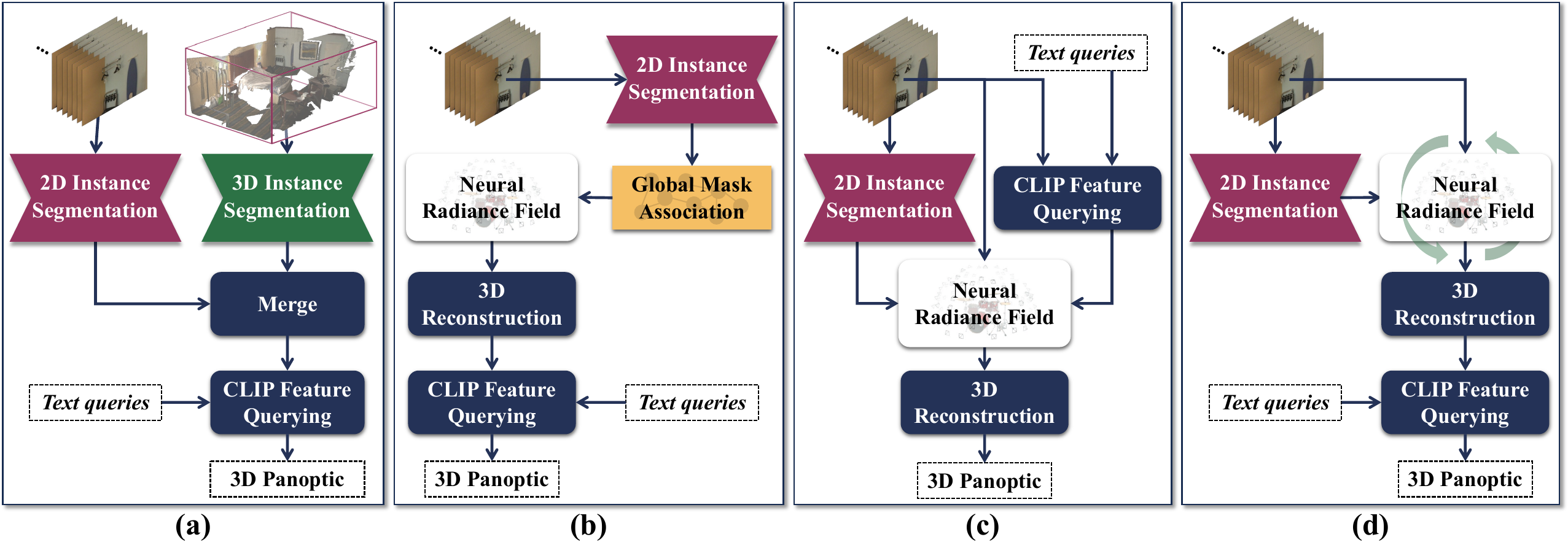}
\caption{\textbf{Mainstream schemes of open-vocabulary 3D panoptic segmentation}.
(a) 2D-3D merging based methods \cite{takmaz2023openmaskd,yin2024sai3d,yan2024maskclustering,nguyen2023open3dis};
(b) 2D image-based methods using mask pre-association \cite{iml,gaussian_grouping};
(c) 2D image-based methods using contrastive loss \cite{pvlff};
(d) our Cues3D.}
\label{fig:rw}
\end{figure}

From the definition above,
it is clear that the 3D panoptic prediction largely depends on the quality of the 3D instance segmentation.
Thus, this paper focuses on enhancing 3D instance segmentation for further improvement on panoptic segmentation.
According to the input modality,
the current open-vocabulary 3D panoptic segmentation methods can be mainly divided into two main groups:
\ding{172} methods merging 2D images and 3D point clouds (shown as Fig. \ref{fig:rw}(a)) and \ding{173} methods based on continuous 2D images (shown as Fig. \ref{fig:rw}(b)(c)).
The former extracts 2D instances and semantic representations from 2D images,
3D masks from 3D point clouds \xf{via point transformers \cite{10249213,9879705,9710703}},
and finally merges both segmentations into a unified 3D representation.
In contrast, the latter typically trains instance/semantic radiance field from 2D instance/semantic prediction via Neural Radiance Fields (NeRF) and reconstructs 3D scene representation containing semantics and instances.


Due to lacking scene-level data, 2D image-based methods require specific techniques to avoid inconsistent 3D instance identity,
i.e., one instance is assigned with several IDs.
To achieve this goal, current methods either learn 3D-consistent instance features through contrastive loss \cite{pvlff} or pre-associate object masks across images \cite{iml}. 
However, compared to 2D-3D merging based approaches,
these methods still perform insufficiently in segmentation accuracy,
and pre-association based on \xf{point matching \cite{10566881,10319695}} may fail when appearing large texture-less areas or similar-appearance objects.

Different from them,
we observe that \textit{the ability of NeRF to encode scene-level appearance and geometry is enough for building a consistent field for 3D instances}.
Motivated by this, we propose \textbf{Cues3D} to exploit NeRF's ability in building instance field.
Specifically,
taking the view-inconsistent instance map obtained from off-the-shelf 2D segmentation models as input,
the NeRF training pipeline is thoroughly re-formulated into three phases.
The first phase is to directly train the NeRF using the view-inconsistent labels,
so that the NeRF initially learns a local-fine but global-inconsistent instance field.
In the second phase,
we design a disambiguation method for 3D instance masks to correct inconsistent instance fields.
After gaining each mask' 3D point cloud by volumetric rendering, 
it separates distinct instances with the same ID and merges consistent ones through 3D matching.
The corrected 3D masks are in turn leveraged to renew the training data of NeRF for cross-view consistency.
The third phase refines the instance field through re-training NeRF.
With the aid of Cues3D,
we obtain highly-consistent 3D instances with high integrity,
which is far better than PVLFF and even better than 2D-3D based SAI3D, as shown in Fig. \ref{fig:visualization_compare}.
Experiments on ScanNet v2 \cite{dai2017scannet}, ScanNet200 \cite{rozenberszki2022language}, Replica \cite{replica19arxiv}, and ScanNet++ \cite{yeshwanthliu2023scannetpp} demonstrate that Cues3D, only using 2D images,
achieves an \textbf{11.1\%} higher AP than SAI3D (a 2D-3D based method) on ScanNet++.
Once using 3D data as an auxiliary,
our method outperforms previous State-of-The-Art (SoTA) 2D-3D based techniques,
surpassing MaskClustering by \textbf{4.9\%} on ScanNet v2 and leading by \textbf{6.9\%} on ScanNet200.
On 3D panoptic/semantic segmentation tasks,
Cues3D still maintains its lead without 3D inputs,
achieving a \textbf{2.3\%} higher mIoU than OpenScene in 3D semantic evaluation,
and more than double ${\text{PQ}}^{\text{scene}}$ of PVLFF in 2D panoptic evaluation on ScanNet v2, and \textbf{1.7\%} higher on Replica.

In summary, the key contribution of our approach lies in:
\begin{itemize}[itemsep=0pt, topsep=0pt]
\item We introduce Cues3D,
a concise and effective zero-shot 3D instance segmentation method exploring NeRF's capabilities in geometry rendering,
without pre-association modules and contrastive loss.
\item We present a carefully designed three-stage training process for NeRF,
which corrects instance inconsistency through rendered 3D information during the training of NeRF. 
\item In extensive experiments on three datasets and three tasks,
Cues3D outperforms previous advanced 2D based approaches, even competes with latest 2D-3D based methods,
while surpasses 2D-3D based methods once using 3D data same as them,
Therefore, Cues3D further promotes the practicality of 3D understanding in embodied AI.
\end{itemize}



\section{Related Work}
\label{sec:relatedwork}

\noindent
\textbf{Novel View Synthesis (NVS)} involves training with a set of posed images and rendering images viewed from new poses.
As a breakthrough,
NeRF \cite{mildenhall2021nerf} has garnered significant attention upon its launch.
Since then, numerous approaches have emerged to enhance NeRF.
To accelerate inference and training of NeRF,
Mip-NeRF \cite{barron2021mip} proposed to render anti-aliased conical frustums instead of rays.
For the same goal,
Instant-NGP \cite{mueller2022instant} greatly accelerated NeRF's training efficiency through hash grid encoding.
To overcome challenges of unbounded scenes,
Mip-NeRF 360 \cite{barron2022mip} proposed non-linear scene parameterization,
online distillation, and a novel distortion-based regularizer.
Recently, 3D Gaussian Splatting (3DGS) \cite{kerbl20233d} broke through the upper limit of NVS.
In this paper,
we employ a well-developed platform NeRFstudio \cite{nerfstudio} instead of 3DGS because it was not released at the beginning of our study.
Fortunately, although Cues3D is built upon NVS,
it is decoupled from the NVS method,
which gives Cues3D the potential to be further improved.

\vspace{5pt}
\noindent
\xf{\textbf{Open-vocabulary Semantic Segmentation} is different from traditional segmentation,
which integrates both visual and text features to enable models to recognize unseen categories during training.
ZegFormer \cite{zegformer} first segments class-agnostic instances,
then extracts image patches of instances to model pixel-level classification as image-level classification.
LSeg \cite{lseg} employs contrastive learning to align pixel features with text features while distancing them from irrelevant ones.
In contrast, OpenSeg \cite{openseg} aligns the segment-level visual features with text features via region-word grounding.
OVSeg \cite{openseg} addresses mask extraction limitations by proposing adaptive mask prompting,
while FC-CLIP \cite{yu2023fcclip} improves efficiency through a single-stage architecture using frozen CLIP features for pixel decoding.
Therefore, we employ FC-CLIP to achieve high efficiency and high accuracy in segmentation.
Furthermore, CLIP \cite{clip} is also utilized to illustrate the lower bound of our method.}

\vspace{5pt}
\noindent
\xf{\textbf{Zero-shot instance segmentation} is the focus of open-vocabulary panoptic segmentation for complementing instance boundaries.
It has achieved breakthrough in recent years via using large-scale datasets.
SAM \cite{sam} achieves remarkable zero-shot results through massive data training,
handling ambiguous prompts via multi-mask prediction and supporting various input modalities (points/boxes). 
However, its computational demands inspire efficient variants:
FastSAM \cite{fastsam} accelerates processing using YOLOv8-Seg with feature pyramids,
MobileSAM \cite{mobilesam} optimizes for mobile devices via Tiny-ViT \cite{tiny_vit} and distillation,
while RobustSAM \cite{chen2024robustsam} introduces degradation-resistant modules.
Other derivatives include EfficientSAM \cite{EfficientSAM}, EdgeSAM \cite{zhou2023edgesam}, and SAM-Lightening \cite{samlightening}.
CropFormer \cite{cropformer} advances segmentation quality by training on high-resolution EntitySeg dataset.
By dividing high-res images into regions for processing,
it achieves sharper boundaries that meet modern high-quality dataset requirements.
Therefore, we employ CropFormer to perform zero-shot instance segmentation in our method.}

\vspace{5pt}
\noindent
\textbf{Open-Vocabulary 3D Scene Understanding} is to locate and recognize 3D elements from scenes based on the open-vocabulary ability of VLMs \cite{clip},
which has received more attention recently compared to closed-set 3D scene understanding methods \cite{qi2017pointnet, jiang2023pointgs, wu2024joint, zhou2023bcinet}.
It is actually divided into several tasks:
\textit{3D semantic} \cite{kerr2023lerf,liu2023weakly,engelmann2024opennerf}, \textit{instance} \cite{takmaz2023openmaskd,wang2023dmnerf}, and \textit{panoptic segmentation} \cite{pvlff}, etc.,
in which we focus on instance and panoptic segmentation.
\textbf{\textit{We review such two tasks together since panoptic segmentation is achieved by projecting semantics onto 3D instances}}.
According to the input modalities,
the existing approaches are divided into two parts:
\ding{172} methods merging 3D point cloud and 2D image, as shown in Fig. \ref{fig:rw} (a),
and \ding{173} methods only taking in 2D images, as shown in Fig. \ref{fig:rw} (b)(c).

As the former methods,
OpenMask3D \cite{takmaz2023openmaskd} proposed the open-vocabulary 3D instance segmentation task, which aggregates per-mask CLIP-based features guided by predicted class-agnostic 3D instance masks.
SAI3D \cite{yin2024sai3d} proposed to synergistically leverage geometric priors and semantic cues derived from Segment Anything Model (SAM).
Open3DIS \cite{nguyen2023open3dis} proposed a 2D-guide-3D Instance Proposal Module to achieve sufficient 3D object binary instance masks.
MaskClustering \cite{yan2024maskclustering} introduced a novel view-consensus rate to assess the relationship between two masks,
boosting the entire 2D-3D instance segmentation.
For the latter type,
PVLFF \cite{pvlff} firstly proposed open-vocabulary 3D panoptic segmentation,
and built implicit instance fields through contrastive learning.
As a \textit{contemporaneous work}, iML \cite{iml} proposed a 2D mask association module as a pre-processing unit to address the view-inconsistency before training NeRF.
Aiming to a similar goal,
Gaussian Grouping \cite{gaussian_grouping} tried to obtain consistent instance labels by tracking but failed to achieve global uniqueness of ID.

Although 2D-3D merging methods generally achieve higher accuracy using both global (3D) and local (2D) data,
they face challenges of capturing high-fidelity 3D point clouds.
2D based methods rely less on inputs but do not perform as well as the former.
Furthermore, the pre-association module might fail inevitably because point matching is infeasible when there are similar-looking objects and large texture-less area.
\xf{In terms of overall architecture,
our Cues3D does not leverage any pre-association modules and contrastive learning, but the 3D knowledge learned by NeRF,
making it streamlined and effective, as shown in Fig. \ref{fig:rw}(d).
In terms of detailed design, Cues3D has some similarities with previous methods.
The design of 3D mask comparison is inspired by the merging method of SAM3D,
which ensures the good efficiency of the comparison step.
In the first stage of Cues3D,
the same training method as DM-NeRF is used to achieve consistency of multi-view instance IDs.}
Actually, in addition to the works above,
there are also many excellent open-vocabulary 3D semantic segmentation methods that are partially related to our method,
such as LERF \cite{kerr2023lerf}, LEGaussain \cite{shi2023language}, LangSplat \cite{qin2023langsplat}, 3D-OVS \cite{liu2023weakly}, OpenNeRF \cite{engelmann2024opennerf}, etc.
Since they do not produce instance-level predictions,
we do not go into more details.

\begin{figure}[t]
\centering
\includegraphics[width=1\textwidth]{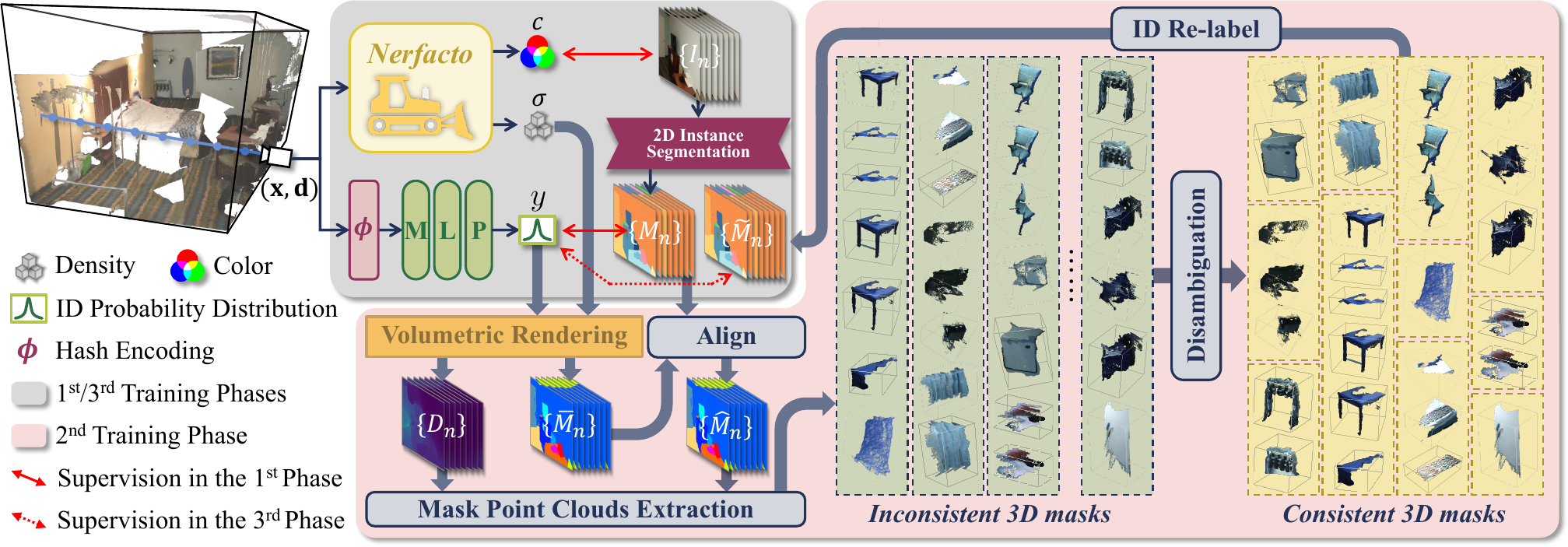}
\caption{\textbf{Structure and training phases of Cues3D}.
The entire training process is divided into three phases.
The first and third phases use RGB images and instance maps to supervise the network.
The second phase re-labels the training data (i.e., instance maps) for the third phase based on the 3D knowledge learned in the first phase.}
\label{fig:pipeline}
\end{figure}

\section{Method}
\label{sec:method}

\subsection{Problem Formulation and Method Overview}

Given a set of continuous posed images $\{I_1, I_2, \dots, I_N\}$,
our method outputs a list of 3D masks for each ``thing'' (instance) and ``stuff'' and their open-vocabulary semantics without the help of depth sensors and reconstructed point clouds.

As illustrated in Fig. \ref{fig:pipeline},
for a ray $r$,
Cues3D takes in the 5D vectors of query point $\mathbf{x}\in\mathbb{R}^3$ and viewing direction $\mathbf{d}\in\mathbb{S}^2$.
The outputs include an RGB color $c\in[0,1]^3$, volume density $\sigma\in[0,\infty)$ and a probability distribution across $U$ instance $y\in [0,1]^{U}$ for $\mathbf{x}$,
where $U$ is a pre-set maximum ID for instances and the $0^\text{th}$ ID indicates empty.
With the definition above,
the inference process can be expressed as:
\begin{equation}
    (c, \sigma, y) = f_\theta(\mathbf{x},\mathbf{d}) 
    \label{eq:1}
\end{equation}
where $\theta$ denotes the parameters of the network.
The network structure of Cues3D has two main modules:
\ding{172} an off-the-shelf NeRFacto \cite{nerfstudio} for learning color $c$ and density $\sigma$,
\ding{173} an instance head for learning the ID of the object the ray hits.
Our instance head has a similar structure to LERF \cite{kerr2023lerf},
containing a hash grid encoding and a multi-layer perceptron (MLP) featuring $U$ output units.

In contrast to previous methods \cite{bhalgat2023contrastive,pvlff,gaussian_grouping},
Cues3D has three training phases,
but keeps the same training iteration to ensure that the time cost remains unchanged.
In particular, we firstly initialize the network for its ability to encode instances (Sec.\ref{sec:s1}),
then the instances learned by the network are disambiguated (Sec.\ref{sec:s2}),
and finally the re-corrected instances are used for refinement (Sec.\ref{sec:s3}).
For open-vocabulary 3D panoptic segmentation,
we give the inference process in Sec.\ref{sec:s4}.



\subsection{Instance Field Initialization for 3D consistency}
\label{sec:s1}
The initial training phase of Cues3D has two goals:
\ding{172} encoding the implicit 3D geometry of the scene and 
\ding{173} roughly recognizing each 3D object without requiring each object to have a unique ID.
The former is easily achieved by the default training strategy of NeRFacto \cite{nerfstudio},
while the latter can be achieved by the Hungarian training scheme \cite{wang2023dmnerf}.

Specifically,
given the images $I_n\in\mathbb{R}^{H\times W\times3}$,
we firstly utilize an off-the-shelf zero-shot 2D segmentation foundation model to produce 2D instance maps $M_n\in\mathbb{Z}_+^{H\times W}$ as the label for training.
Secondly, on each image $I_n$,
we sample a set of rays $\{r^1_n,r^2_n,\dots,r^J_n\}$ and predict these rays' instance $\{y^1_n,y^2_n,\dots,y^J_n\}$ through the network $f_\theta(\cdot)$ in equation \ref{eq:1}.
Referring to DM-NeRF \cite{wang2023dmnerf},
the soft intersection-over-union (sIoU) and cross-entropy (CE) cost \cite{yang2019learning} are used to align the prediction $\{y^j_n|j\in[1,J]\}$ with the label $M_n$.
\begin{align}
&\mathcal{C}^{u,t}_{sIoU} = 
\frac{-\sum_{j=1}^J \big[ y^j_n(u) \!\times\! \mathbf{M}_{n}^{j,t}\big]}{\sum_{j=1}^J y^j_n(u) \!+\! \sum_{j=1}^J \mathbf{M}_{n}^{j,t} \!-\! \sum_{j=1}^J \big[ y^j_n(u) \!\times\! \mathbf{M}_{n}^{j,t}\big]}\\
&\mathcal{C}^{u,t}_{CE} = 
-\frac{1}{J} \sum\nolimits_{j=1}^J \Big[\mathbf{M}_{n}^{j,t} \times\log\big(y^j_n(u)\big) + (1-\mathbf{M}_{n}^{j,t})\times\log\big(1-y^j_n(u)\big)\Big]
\end{align}
where $\mathbf{M}_n\!\in\!\{0,1\}^{H\times W\times T_n}$ denotes the one-hot version of $M_n$,
and $T_n$ is the maximum ID in $M_n$.
$\mathbf{M}_{n}^{j,t} \!=\! \mathbf{M}_n\big(\mathtt{pix}(r^j_n),t\big)$ represents the $t^\text{th}$ channel value of the pixel projected from $r^j_n$ on map $M_n$.
$u$ denotes an ID inside the pre-set instance list $[1,U]$,
and $t\in[1,T_n]$ denotes an instance ID on the label $M_n$.
Then, the cost $(\mathcal{C}^{u,t}_{sIoU} + \mathcal{C}^{u,t}_{CE})$ is fed into Hungarian algorithm\cite{kuhn1955hungarian} to optimally associate a predicted ID $u$ to an ID $t$ on $M_n$,
which allows us to reorder the IDs in $y^j_n$ to align with $T_i$ instances in label $M_n$.
Finally, the IDs of rays on one image are supervised as follows:
\begin{equation}
\mathcal{L}_{1st} = \frac{1}{T_n}\sum\nolimits_{t=1}^{T_n} (\mathcal{C}^{t,t}_{sIoU} + \mathcal{C}^{t,t}_{CE})
\end{equation}
In each training iteration,
since the instance IDs of rays are aligned to $M_n$,
each point in the 3D implicit field is supervised to obtain a fixed ID,
i.e., one object in multiple views has the same ID.
However, multiple objects may be ambiguously assigned to the same ID.

\subsection{3D Instance Identity Disambiguation for Global ID Uniqueness}
\label{sec:s2}
The second phase aims to achieve global ID uniqueness for each instance using 3D geometry learned in the first phase.
Differentiating objects with the same ID from local views is difficult,
but easier from a global 3D view.
Motivated by this,
we first extract the 3D mask of each rough instance and then disambiguate their IDs in a global 3D view.

\subsubsection{NeRF-based 3D Masks Extraction}
For all images $\{I_n|n\!\in\![1,N]\}$,
the point clouds of all masks are extracted in preparation for disambiguation.
Specifically,
for each ray $\{r^j_n|j\!\in\![1,H\!\times\! W]\}$ on an image $I_n$,
we first predict their volume densities $\{\sigma^j_n\}$ and instance IDs $\{\arg \max_{u} y^j_n(u)\}$.
Taking them as inputs,
volumetric rendering \cite{kajiya1984ray} is then employed to synthesize the depth map $D_n$ and 2D instance map $\Bar{M}_n$ ($\Bar{M}_n\!\in\!\mathbb{R}^{H\times W}$).
Notably,
we observe the complementarities of the synthesized instance map $\Bar{M}_n$ and the label $M_n$:
\begin{itemize}[itemsep=0pt, topsep=0pt, parsep=0pt, partopsep=0pt]
\item $\bar{M}_n$ holds consistent IDs across views but suffers from inaccurate mask.
\item $M_n$ has accurate masks but with inconsistent IDs across views.
\end{itemize}
\begin{figure}[h]
\centering
\includegraphics[width=1\textwidth]{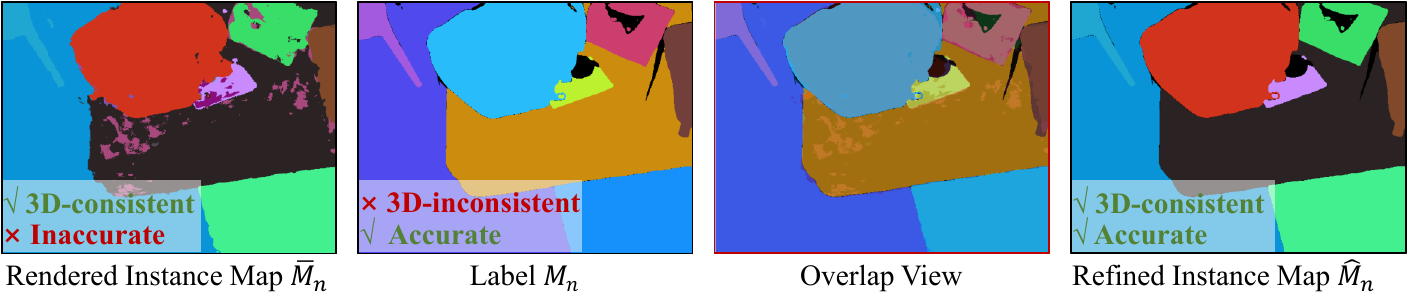}
\caption{\textbf{Visualization of instance maps $\bar{M}_n$, $M_n$, $\hat{M}_n$.}}
\vspace{-10pt}
\label{fig:merge}
\end{figure}
Therefore, we combine both maps to achieve accurate and 3D-consistent masks.
In detail, each mask in $M_n$ is filled with the instance ID of mask in $\Bar{M}_n$ with the largest overlap,
which generates a new instance map $\hat{M}_n\in\mathbb{R}^{H\times W}$.
\begin{equation}
\hat{M}_n(p) = \arg\max_u \big(\big|\mathds{1}[\Bar{M}_n==u] \cap \mathds{1}[M_n==t]\big|\big) \quad \forall p,M_n(p)==t
\label{eq:iou}
\end{equation}
where $\mathds{1}[\cdot]$ is an indicator function that equals 1 if the inner element is true and 0 otherwise.
We traverse all instance IDs $t\in[1,T_n]$ to generate $\hat{M}_n$.

Subsequently,
we extract each instance's depth from the rendered depth map $D_n$
and transform them into 3D point clouds.
All the point clouds are in the same world coordinate system maintained by NeRFacto.
In this way, we process all images $\{I_1,I_2,...,I_N\}$ and capture numerous 3D instance masks grouped by rendered instance ID,
denoted as $\{\mathbf{G}_1, \mathbf{G}_2, ..., \mathbf{G}_{\Bar{U}} \}$,
where $\Bar{U} \leq U$ is the maximum instance ID.
The groups are represented in detail:
\begin{itemize}[itemsep=0pt, topsep=0pt, parsep=0pt, partopsep=0pt]
    \item Each group, i.e., $\mathbf{G}_u$, contains a set of 3D masks $\{P_1, P_2,..., P_{L_u}\}$.
    \item Each mask, i.e., $P_l$ ($l\in[1,L_u]$), consists of $Q_l$ 3D points: $P_l\in\mathbb{R}^{Q_l \times 4}$.
    \item Each point is defined by a 3D coordinate and the mask index, $(\mathrm{x},\mathrm{y},\mathrm{z},l)$.
\end{itemize}
In addition, each mask $P_l$ can be found through a hierarchical index $Z_l = \{(u,n)\}$,
which means that $P_l$ represents the instance $u$ in the map $\hat{M}_n$.
In this way,
we define an indices group $\mathbf{Z}_u = \{Z_1,Z_2,...,Z_{L_u}\}$ for $\mathbf{G}_u$.
Such indices are used as a lookup table in the following steps to record distinct objects with the same ID.
\xf{Since each point cloud set $\mathbf{G}_u$ is processed by the same instance disambiguation operation,
we omit the symbol $u$ below to simplify the description.}

\begin{figure}[t]
\centering
\includegraphics[width=1\textwidth]{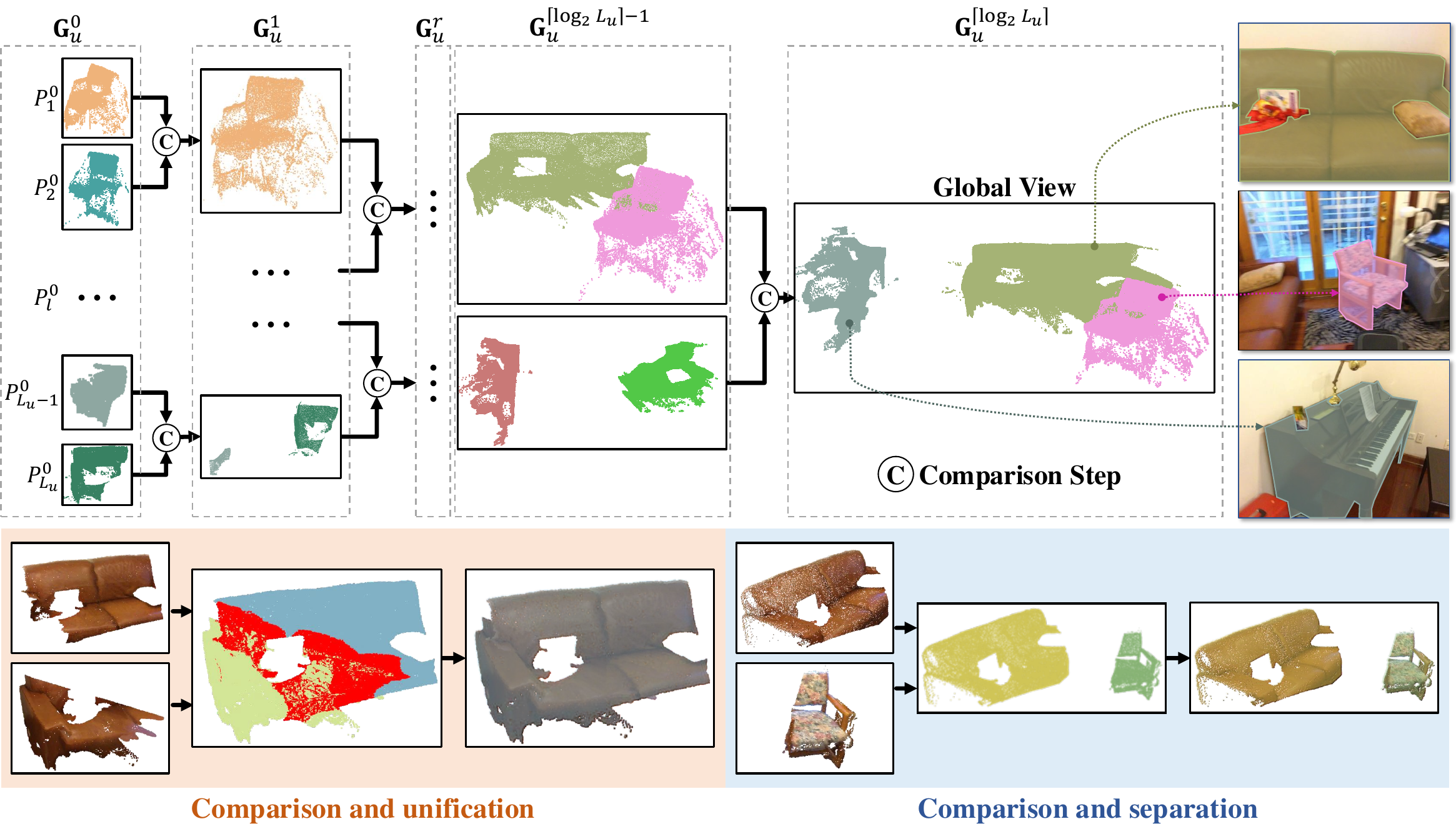}
\caption{\textbf{Pipeline of Instance Disambiguation.}
The red marks the 3D point pairs matched successfully,
while other colors indicate the labels of 3D points.}
\vspace{-10pt}
\label{fig:disambi}
\end{figure}

\subsubsection{Instance Disambiguation}
\label{sec:disambiguation}
Given a group of 3D masks $\mathbf{G}$ and their indices $\mathbf{Z}$,
the disambiguation step aims to compare two 3D masks inside $\mathbf{G}$ gradually,
which unifies high-overlap objects into the same ID and separates low-overlap objects into distinct IDs.

As illustrated in Fig. \ref{fig:disambi} and Algorithm. \ref{alg:disambig},
the disambiguation follows a hierarchical structure,
which contains multiple rounds of pair-wise comparison.
Specifically,
all 3D masks in $\mathbf{G}$ are taken as the elements to be compared in the initial round,
denoted as $\mathbf{G}^\mathit{0} = \{P^\mathit{0}_1, P^\mathit{0}_2,..., P^\mathit{0}_{L}\}$.
In each round,
we compare every two adjacent 3D masks, e.g., $P^\mathtt{r}_{2l-1}$ and $P^\mathtt{r}_{2l}$,
and combine them to $P^\mathtt{r+1}_{l}$ for next round,
where $\mathtt{r}$ is round number.
\begin{equation}
P^\mathtt{r+1}_{l} \gets P^\mathtt{r}_{2l - 1} \cup P^\mathtt{r}_{2l}; \quad
\mathbf{G}^{\mathtt{r}+1} := \mathbf{G}^{\mathtt{r}+1} \cup \{P^\mathtt{r+1}_{l}\}
\end{equation}
The comparison would be terminated until all 3D masks are combined into a global view.
In the following we give a detailed way of single comparison.

\vspace{6pt}
\noindent
\textbf{3D Masks Matching in A Unified View.}
Since all the 3D masks are in the same coordinate system maintained by NeRFacto,
we establish voxel-based nearest neighbor matching between 3D masks.
\xf{The number of points in a point cloud affects the matching speed.
To reduce computational complexity,
we first perform voxelization on each point cloud $P_l$,
which reduces the point number in $P_l$.}
Taking two adjacent 3D masks $P^\mathtt{r}_{2l-1}$ and $P^\mathtt{r}_{2l}$ as examples,
we then compute the nearest neighbor point pairs whose distance is less than $\tau_\mathbf{d}$,
denoted as $P^\mathtt{r}_{2l-1,2l}=\{(\mathbf{p}^{m_1}_{2l-1}, \mathbf{p}^{m_2}_{2l})|m_1 \leq |P^\mathtt{r}_{2l-1}|, m_2 \leq |P^\mathtt{r}_{2l}|\}$,
shown as the red points in the cases of Fig. \ref{fig:disambi},
\xf{where $\tau_\mathbf{d}$ is the maximum distance between two matched points}.
Then, taking the $4^\text{th}$ number of each point as its instance label,
we count the overlap between each two labels in this matching:
$O^\mathtt{r}_{2l-1,2l}=\{(\mathtt{l}_a,\mathtt{l}_b):o_{a,b}\}$,
where $O^\mathtt{r}_{2l-1,2l}$ is a dictionary,
$\mathtt{l}_a$ and $\mathtt{l}_b$ are the labels of points,
and $o_{a,b}$ denotes the number of points matched between the labels $\mathtt{l}_a$ and $\mathtt{l}_b$.
This overlap indicates that two masks with different labels are or are not the same object.



\begin{algorithm}[h]
\scriptsize
\caption{Instance Disambiguation}
\label{algorithm}
\KwIn{The 3D masks group $\mathbf{G}^\mathit{0}$ = \{$P^\mathit{0}_1$, $P^\mathit{0}_2$, ..., $P^\mathit{0}_{L}$\},
indices $\mathbf{Z}$ = \{$Z_1$, $Z_2$, ..., $Z_{L}$\} }
\KwOut{Indices of corrected 3D masks $\mathbf{\bar{Z}}$ =  \{$Z_1$ $Z_2$ ... $Z_{\bar{L}}$\}}
Initializing the round of disambiguation: $\mathtt{r} \gets 0$\\
\While{$|\mathbf{G}^{\mathtt{r}}| > 1$}{
    Creating an empty set: $\mathbf{G}^{\mathtt{r}+1} \gets \emptyset$\\
    \ForEach{$l$ \textbf{in} $\{1, 2, ..., |\mathbf{G}^\mathtt{r}|/2\}$}{
        Matching $P_{2l - 1}^{\mathtt{r}}$ and $P_{2l}^{\mathtt{r}}$\\
        Saving all matched point pairs:
        $P^\mathtt{r}_{2l-1,2l}=\{(\mathbf{p}^{m_1}_{2l-1}, \mathbf{p}^{m_2}_{2l})|m_1 \leq |P^\mathtt{r}_{2l-1}|, m_2 \leq |P^\mathtt{r}_{2l}|\}$\\
        Creating an empty directory: $O^\mathtt{r}_{2l-1,2l}\gets \emptyset$\\
        \ForEach{$(\mathbf{p}^{m_1}_{2l-1},\mathbf{p}^{m_2}_{2l})$ \textbf{in} $P^\mathtt{r}_{2l-1,2l}$}{
                \eIf{$(\mathbf{p}^{m_1}_{2l-1}[4],\mathbf{p}^{m_2}_{2l}[4]) \in O^\mathtt{r}_{2l-1,2l}.key$}
                {
                $O^\mathtt{r}_{2l-1,2l}[(\mathbf{p}^{m_1}_{2l-1}[4], \mathbf{p}^{m_2}_{2l}[4])] := O^\mathtt{r}_{2l-1,2l}[(\mathbf{p}^{m_1}_{2l-1}[4], \mathbf{p}^{m_2}_{2l}[4])] + 1$
                }
                {
                $O^\mathtt{r}_{2l-1,2l}[(\mathbf{p}^{m_1}_{2l-1}[4], \mathbf{p}^{m_2}_{2l}[4])] := 1$
                }
            }
        \ForEach{$(\mathtt{l}_a,\mathtt{l}_b):o_{a,b}$ \textbf{in} $O^\mathtt{r}_{2l-1,2l}$}{
            $less\_point\_in\_two = \min\{|P_{2l-1}^{\mathtt{r}}[:,4]==\mathtt{l}_a|, |P_{2l}^{\mathtt{r}}[:,4]==\mathtt{l}_b|\}$\\
            \If{$o_{a,b} > less\_point\_in\_two/2$ \textbf{or} $o_{a,b} > \tau_\mathbf{n}$}{
                Update $P_{2l}^{\mathtt{r}}[:,4]$ that are equal to $\mathtt{l}_b$ to $\mathtt{l}_a$\\
                $Z_{\mathtt{l}_a} := Z_{\mathtt{l}_a} \cup Z_{\mathtt{l}_b}; \quad \mathbf{Z} := \mathbf{Z} / \{Z_{\mathtt{l}_b}\}$
            }
        }
        $P^\mathtt{r+1}_{l} \gets P^\mathtt{r}_{2l - 1} \cup P^\mathtt{r}_{2l}$; \quad
        $\mathbf{G}^{\mathtt{r}+1} := \mathbf{G}^{\mathtt{r}+1} \cup \{P^\mathtt{r+1}_{l}\}$\\
    }
    $\mathtt{r} := \mathtt{r} + 1$
}
$\mathbf{\bar{Z}} \gets \mathbf{Z}$
\label{alg:disambig}
\end{algorithm}

\vspace{6pt}
\noindent
\textbf{Updating Indices of 3D Masks and Labels of 3D Points.}
According to the matched points,
we subsequently update the indices of the 3D masks.
In more detail,
for each pair of labels,
e.g. $(\mathtt{l}_a,\mathtt{l}_b)$,
in $O^\mathtt{r}_{2l-1,2l}$,
we set two conditions referred to SAM3D:
\begin{itemize}[itemsep=0pt, topsep=0pt, parsep=0pt, partopsep=0pt]
\item The matched points $o_{a,b}$ is more than half number of the smaller one.
\item The number of matched points $o_{a,b}$ is greater than $\tau_\mathbf{n}$.
\end{itemize}
\xf{The purpose of setting $\tau_\mathbf{n}$ is that for large objects,
such as beds, pianos, etc.,
the point clouds between the two views may lack enough overlap area beyond half of the smallest object.
Therefore, $\tau_\mathbf{n}$ is set as the lower limit of this overlap to ensure that the two point clouds of large objects can be matched correctly.}
If one of the conditions is true,
the points of the two labels can be considered to belong to the same object.
Thus we combine their indices into $Z_{\mathtt{l}_a}$:
\begin{equation}
Z_{\mathtt{l}_a} := Z_{\mathtt{l}_a} \cup Z_{\mathtt{l}_b}; \quad
\mathbf{Z} := \mathbf{Z} / \{Z_{\mathtt{l}_b}\}
\end{equation}
And all 3D points with label $\mathtt{l}_b$ are assigned as $\mathtt{l}_a$:
\begin{equation}
P^\mathtt{r}_{2l - 1}[i, 4] := \begin{cases} 
\mathtt{l}_b & \text{if } P^\mathtt{r}_{2l - 1}[i, 4] = \mathtt{l}_b \\
P^\mathtt{r}_{2l - 1}[i, 4] & \text{otherwise}
\end{cases}
\end{equation}
Otherwise, these points are considered to belong to two distinct objects.

Through such a process above,
we compare two point clouds $P^\mathtt{r}_{2l-1}$ and $P^\mathtt{r}_{2l}$,
unify the ID of masks belonging to the same object,
and merge their indices.
In this way, the 3D Mask indices of the same object are gradually gathered,
finally resulting in the indices in the last round:
$\bar{\mathbf{Z}} = \{Z_1,Z_2,...,Z_{\bar{L}}\}$,
where $\bar{L}$ is the number of true objects in $\mathbf{G}$,
and each element $Z_{\bar{l}}$ ($ \bar{l} \in [ 1, \bar{L} ]$) contains the mask indices of an instance with global uniqueness.
\xf{The process of instance disambiguation can be viewed in more detail from a flowchart in Section \ref{sec:flow}.}

\begin{figure}[t]
\centering
\includegraphics[width=1\textwidth]{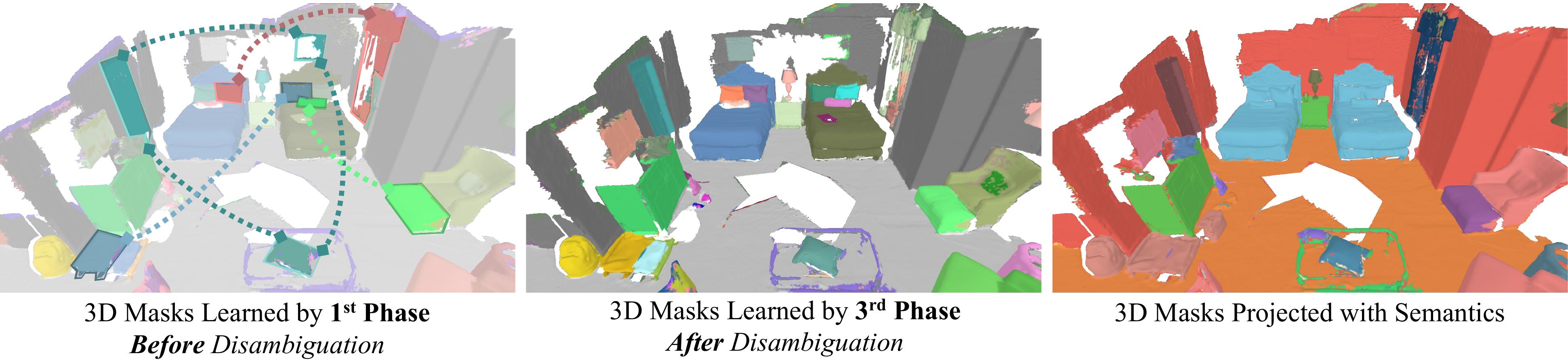}
\caption{\textbf{Visualization of the 3D masks} before/after disambiguation and voting-based semantic projection.
Several distinct objects with the same IDs are highlighted in the leftmost image.}
\label{fig:compare_phases}
\end{figure}

\vspace{6pt}
\noindent
\textbf{3D Masks ID Correction After Comparison.}
According to the refined mask indices $\bar{\mathbf{Z}}$,
the instance IDs on the maps $\{M_n\}$ can be easily replaced by each object's truly unique instance ID,
which forms the corrected segmentation maps $\{\Tilde{M}_n\}$.
Fig. \ref{fig:compare_phases} shows the 3D masks before and after disambiguation.
Before disambiguation,
multiple objects have the same ID,
such as the ``\textit{curtains}'' and ``\textit{pillows}'',
while each object has an isolated ID after disambiguation.

\subsection{Instance Field Fine-Tuning}
\label{sec:s3}

In the third phase,
we directly supervise each ray’s probability distribution using Cross-Entropy loss,
ensuring that the NeRF model obtains a consistent instance field.
Specifically,
to reduce the time cost of training,
we first randomly sample a batch of rays $R=\{y_j|j\in[1,\mathbf{J}],y_j\in\mathbb{R}^{U}\}$ from all the $N$ images,
which is in contrast to the first phase that samples rays from each image,
Then, we search the corrected instance ID of each ray $\{\Tilde{y}_j|j\in[1,\mathbf{J}],\Tilde{y}_j\in\mathbb{R}^{U}\}$ from the corrected instance map $\{\Tilde{M}_n|n\in[1,N]\}$.
The cross-entropy loss of the third phase is formulated as follows:
\begin{equation}
\mathcal{L}_{3^\text{rd}} = - \frac{1}{\mathbf{J}} \sum\nolimits_{j=1}^{\mathbf{J}} \frac{1}{U} \sum\nolimits_{u=1}^{U} \Big[ \Tilde{y}_{j}(u) \times \log \big(y_j(u) \big) \Big]
\end{equation}

\subsection{Visual-Language Semantic Voting across Multiple Views}
\label{sec:s4}
The consistent instances in Cues3D allow the application of open-vocabulary queries for 3D objects,
which can be used in many downstream tasks.
To this end, we back-project semantic features onto each instance through voting.
Specifically,
we first extract the semantic feature $F_n$ and 2D semantic segmentation $S_n$ using VLMs,
e.g., CLIP \cite{clip}, OpenSeg \cite{ghiasi2021open}, and FC-CLIP \cite{yu2023fcclip}.
Secondly,
we process each correct instance map $\Tilde{M}_n$.
For each map $\Tilde{M}_n\in\mathbb{R}^{H\times W}$,
we construct an instance-semantic voting matrix $V_n\in\mathbb{Z}_+^{U\times C}$.
During building $V_n$,
each instance finds its corresponding semantic class $\hat{c} = \arg\max_\mathbf{c}\mathcal{I}^{u,\mathbf{c}}(\Tilde{M}_n,S_n)$
(according to Eq. \ref{eq:iou}),
and $V_n(u,\hat{c})$ is set to 1,
where $\mathcal{I}^{u,\mathbf{c}}(\Tilde{M}_n,S_n)$ denotes the intersection between the $u^\text{th}$ instance on $\Tilde{M}_n$ and the $\mathbf{c}^\text{th}$ region on $S_n$.
$u\in[1,\bar{U}]$ denotes an instance ID,
and $\hat{c}\in[1,C]$ denotes one of the semantic classes.
All other numbers in $V_n$ that are not 1 are set to 0.
Third, we sum all voting matrices of a single image: $V = \sum_{n\in[1,N]}V_n$,
which is taken as the final voting matrix between instances and semantic classes.
In this matrix,
each instance obtains a semantic class with the most votes.
Finally, during the inference,
each 3D instance rendered by NeRF can obtain a suitable semantic class according to the matrix $V$.
As shown in Fig .\ref{fig:compare_phases},
with the fine-grained 3D instance,
the open-vocabulary semantics are accurately attached to an entire object without overflowing.
Since we have obtained multi-view consistent instances after the training phase,
this voting method effectively avoids semantics inconsistency across views.
Another advantage is that combining the multi-view votes makes the semantic projection more reliable.

\section{Experiments}
\label{sec:exp}

\subsection{Experiment Setting}


\paragraph{Implementation Details}

For zero-shot 2D instance segmentation,
CropFormer \cite{cropformer} is employed to obtain $M_n$ in Sec. \ref{sec:s1}.
For a comprehensive evaluation,
we assess Cues3D adopted with several open-vocabulary semantics extraction methods in Sec. \ref{sec:s4},
i.e., CLIP \cite{clip},
OpenSeg\cite{ghiasi2021open},
and FC-CLIP\cite{yu2023fcclip}.
Note that,
when using CLIP \cite{clip} in semantic voting,
we exclude masks with a similarity lower than 0.7 to reduce noise interference.
In terms of training setting,
we employ a total of 30,000 training iterations,
which is same as the original NeRFacto \cite{nerfstudio},
the first 20,000 for the first phase and the last 10,000 for the third phase.
The maximum number of instances, i.e., $U$, is set to 200,
and $\tau_\mathbf{d}$ = 0.075 and $\tau_\mathbf{n}$ = 50 for instance disambiguation.
\xf{The voxel size in the voxelization operation is set to 0.05.}
The MLP used for learning instance features has 4 hidden layers with width 256 before the final $U$-dimension instance output.
All images of each dataset are scaled to $640\times480$.
For each scene,
we sample 80\% of images from the image sequence at equal intervals for the training set and used the remaining 20\% as the test set.


\begin{table}
\centering
        \resizebox{0.6\linewidth}{!}{
        \begin{tabular}{l|ccc}
        \toprule
        \textbf{3D Tasks}   & \textbf{ScanNet} \cite{dai2017scannet}  & \textbf{ScanNet++} \cite{yeshwanthliu2023scannetpp} & \textbf{Replica} \cite{replica19arxiv} \\
        \midrule
        \textbf{Panoptic}   & $\mathbf{2D}$/\textcolor{gray}{$\mathit{3D}$}    & \textcolor{gray}{$\mathit{3D}$} & $\mathbf{2D}$ \\
        \textbf{Instance}   & $\mathbf{3D}$ & $\mathbf{3D}$          &  -            \\
        \textbf{Semantic}   & $\mathbf{2D/3D}$ &  -                     & $\mathbf{2D}$ \\
        \bottomrule
        \multicolumn{4}{l}{\footnotesize $\mathbf{2D}$: evaluating 2D metrics on 2D images.}\\
        \multicolumn{4}{l}{\footnotesize $\mathbf{3D}$: evaluating 3D metrics in the 3D space.}\\
        \multicolumn{4}{l}{\footnotesize \textcolor{gray}{$\mathit{3D}$}: omitting 3D Panoptic evaluation due to instance evaluation.}
        \end{tabular}}
    \caption{\textbf{Evaluation datasets and tasks}.
    ScanNet 200 and ScanNet v2 are called ScanNet for short.}
    \label{table:benchmark}
\end{table}

\paragraph{Datasets}
We employ four datasets,
i.e., ScanNet v2\cite{dai2017scannet},
ScanNet200\cite{rozenberszki2022language},
Replica\cite{replica19arxiv},
and ScanNet++\cite{yeshwanthliu2023scannetpp}, for evaluation.
As a widely used dataset,
ScanNet v2 \cite{dai2017scannet} provides many 3D scenes containing 3D point clouds,
2D images, depth images,
and fine-grained 2D/3D instances and semantic annotations,
while ScanNet200 \cite{rozenberszki2022language} extend it to more classes and annotations.
Replica \cite{replica19arxiv} is also commonly used in NVS methods.
It provides high-quality reconstructions of various indoor spaces.
ScanNet++ \cite{yeshwanthliu2023scannetpp} is a recently proposed dataset with comprehensive upgrades in image quality, annotation accuracy and quantity compared to ScanNet v2.
Table \ref{table:benchmark} shows our detailed experimental setting.
Following the experiment setting of other 2D-image based methods \cite{2023_CVPR_panopticlifting,iml},
we employ 12 scenes for ScanNet v2/200, 6 scenes from Replica and 8 scenes from ScanNet++ for evaluation.
\textit{Note that it is common practice to conduct experiments on select scenes rather than the entire dataset,
due to the prohibitive time cost of training on the full dataset.}
Table \ref{table:dataset_scenes} shows the scenes we used.

\begin{table}[t]
\scriptsize
\centering
\begin{tabular}{l|ccc}
\toprule
\textbf{Datasets} & \multicolumn{3}{c}{\textbf{Scene Names}}   \\
\midrule
\multirow{4}{*}{\textbf{ScanNet} \cite{dai2017scannet}} & scene0050\_02 & scene0144\_01 & scene0221\_01 \\
        & scene0300\_01 & scene0354\_00 & scene0389\_00 \\
        & scene0423\_02 & scene0427\_00 & scene0494\_00 \\
        & scene0616\_00 & scene0645\_02 & scene0693\_00 \\
\midrule
\multirow{2}{*}{\textbf{Replica} \cite{replica19arxiv}} & office\_0 & office\_2 & office\_4 \\
        & room\_0 & room\_1 & room\_2       \\ \midrule
\multirow{3}{*}{\textbf{ScanNet++} \cite{yeshwanthliu2023scannetpp}} & 7b6477cb95 & 825d228aec & a24f64f7fb\\
        & 1ada7a0617 & 5748ce6f01 & 40aec5fffa \\
        & bcd2436daf & 6115eddb86 \\
\bottomrule
\end{tabular}
\caption{\textbf{Datasets and scene names used for evaluation}.}
\label{table:dataset_scenes}
\end{table}

\paragraph{Evaluation Metrics}
For 3D \textbf{instance} segmentation,
we ignore semantic labels and only assess the quality of the 2D and 3D masks,
with the standard average precision (AP) at 25\% and 50\% IoU and the mean AP (mAP) from 50\% to 95\% at 5\% intervals.
For 3D \textbf{panoptic} segmentation,
we adopt the mIoU and scene-level panoptic quality ($\text{PQ}^\text{scene}$) in the 2D space proposed in \cite{2023_CVPR_panopticlifting}. 
For 3D \textbf{semantic} segmentation,
we evaluate the mean IoU (mIoU) and mean accuracy (mAcc) of 2D/3D semantic prediction.
When comparing NVS methods on open-vocabulary 3D semantic segmentation,
we use all categories from the 2D annotations within each scene as prompts.
While for 2D-3D merging based methods like OpenScene\cite{peng2023openscene},
we adopt 20 categories from ScanNet v2 to maintain consistency.
Note that,
for comparison with 2D-3D based methods,
we project the 2D prediction of image-based methods to the reconstructed point cloud when computing 3D metrics.

\begin{table}
\centering
\resizebox{1\linewidth}{!}{
\begin{tabular}{l l ccc ccc ccc}
\toprule
\multirow{2.5}{*}{\centering \textbf{Method}} & \multirow{2.5}{*}{\centering \textbf{Inputs}} & \multicolumn{3}{c}{\textbf{ScanNet v2}} & \multicolumn{3}{c}{\textbf{ScanNet200}} & \multicolumn{3}{c}{\textbf{ScanNet++}} \\
\cmidrule(r){3-5} \cmidrule(r){6-8} \cmidrule(r){9-11}
& & \textbf{AP} & \textbf{AP50} & \textbf{AP25} & \textbf{AP} & \textbf{AP50} & \textbf{AP25} & \textbf{AP} & \textbf{AP50} & \textbf{AP25} \\
\midrule
GaussianGrouping \cite{gaussian_grouping}   & $\mathcal{R}$    & 04.4 & 09.3 & 34.3 & 03.3 & 09.0 & 34.3 & - & - & - \\
\textbf{Cues3D}                                   & $\mathcal{R}$    & \textbf{34.7} & \textbf{57.9} & \textbf{75.7} & \textbf{34.9} & \textbf{61.2} & \textbf{76.9} & \textbf{28.1} & \textbf{46.1} & \textbf{61.4} \\
\midrule
SAM3D \cite{yang2023sam3d}                  & $\mathcal{R}$+$\mathcal{D}$+$\mathcal{P}$ & 14.3 & 31.6 & 53.7 & 12.8 & 28.4 & 51.1 & 11.6 & 26.3 & 45.3 \\
SAI3D \cite{yin2024sai3d}                   & $\mathcal{R}$+$\mathcal{D}$+$\mathcal{P}$ & 34.9 & 52.6 & 69.7 & 34.2 & 50.5 & 66.6 & 17.0 & 30.6 & 42.9 \\
Open3DIS (w/o 3DNet)\cite{nguyen2023open3dis}& $\mathcal{R}$+$\mathcal{D}$+$\mathcal{P}$    & 28.8 & 45.7 & 66.3 & 29.7 & 48.3 & 68.0 & 24.8 & 40.3 & 49.9 \\
Open3DIS \cite{nguyen2023open3dis}          & $\mathcal{R}$+$\mathcal{D}$+$\mathcal{P}$ & \underline{42.8} & 59.9 & 77.5 & \underline{41.1} & 58.1 & 76.1 & 24.6 & 38.1 & 46.6 \\
MaskClustering \cite{yan2024maskclustering} & $\mathcal{R}$+$\mathcal{D}$+$\mathcal{P}$ & 40.9 & \textbf{64.3} & \textbf{78.6} & \underline{41.1} & \underline{65.3} & \textbf{78.1} & \textbf{34.9} & \textbf{53.1} & \textbf{66.7} \\
\textbf{Cues3D$+$Superpoint}                      & $\mathcal{R}$+$\mathcal{D}$+$\mathcal{P}$ & \textbf{45.8} & \underline{63.3} & \underline{78.2} & \textbf{48.0} & \textbf{66.4} & \underline{77.7} & \underline{34.4} & \underline{51.7} & \underline{62.4} \\
\bottomrule
\end{tabular}}
\caption{\textbf{3D evaluation of class-agnostic 3D instance segmentation}.
``$\mathcal{R}$'' denotes using RGB images,
``$\mathcal{D}$'' for metric depth images,
and ``$\mathcal{P}$'' for reconstructed point clouds.
Best results are in bold, second best are underlined.}
\label{table:instance}
\end{table}

\subsection{Comparison to SoTA methods}
\label{exp:sota}
\paragraph{Results of 3D Instance Segmentation}
We primarily compare Cues3D with 2D-3D based methods:
SAM3D \cite{yang2023sam3d},
SAI3D\cite{yin2024sai3d},
Open3DIS\cite{nguyen2023open3dis}
and MaskClustering\cite{yan2024maskclustering},
while excluding iML \cite{iml} due to lacking code and details for reproduction \footnote{iML did not provide the data-splitting file for training and test on ScanNet and Replica.}.
Table \ref{table:instance} reports the performance of ScanNet v2, ScanNet200 and ScanNet++.
When using only 2D images as input,
Cues3D comprehensively outperforms SAM3D and surpasses the latest SAI3D by +11.1\% AP on ScanNet++,
despite not being higher than MaskClustering (\textit{A contemporaneous work}).
When we use 3D superpoint as an auxiliary like other 2D-3D merging methods,
Cues3D is boosted by +11.1\% AP on ScanNet v2, +13.1\% AP on ScanNet200, and +6.3\% on ScanNet++,
and surpasses the majority of 2D-3D merging based methods.
In addition,
we found that GaussianGrouping \cite{gaussian_grouping} performs poorly on these dataset,
which is due to its object tracking algorithm for associating objects in image sequences.
The short-term object tracking algorithms cannot re-detect objects when objects disappear and then reappear,
resulting in incorrect pseudo-labels for supervision.



\begin{table}[t]
\centering
\resizebox{0.6\linewidth}{!}{
\begin{tabular}{l c c c c c c}
\toprule
\multirow{2.5}{*}{\textbf{Method}} & \multirow{2.5}{*}{\begin{tabular}{c}\textbf{Image}\\\textbf{Feature}\end{tabular}} & \multicolumn{2}{c}{\textbf{ScanNet v2}} & \multicolumn{2}{c}{\textbf{Replica}} \\
\cmidrule(r){3-4} \cmidrule(r){5-6}
& & \textbf{PQ}$^{\textbf{scene}}$ & \textbf{mIoU} & \textbf{PQ}$^{\textbf{scene}}$ & \textbf{mIoU} \\
\midrule
PVLFF \cite{pvlff} & LSeg & 13.2 & \underline{39.7} & 11.0 & 26.2 \\
\textbf{Cues3D} & CLIP & \underline{30.5} & 38.5 & \underline{12.7} & \underline{27.3} \\
\textbf{Cues3D} & FC-CLIP & \textbf{37.2} & \textbf{56.7} & \textbf{16.2} & \textbf{41.9} \\
\bottomrule
\end{tabular}}
\caption{\textbf{2D evaluation of open-vocabulary 3D panoptic segmentation} based on NVS.
Best results are in bold, second best are underlined.}
\label{table:panoptic}
\end{table}

\paragraph{Results of 3D Panoptic Segmentation}
Since the comparison of 3D panoptic segmentation is consistent with that of 3D instance segmentation when using the same VLM,
we only evaluate current NVS methods of open-vocabulary 3D panoptic segmentation.
As shown in Table \ref{table:panoptic},
compared with PVLFF using LSeg\cite{lseg},
we utilize the earliest CLIP\cite{clip} but still achieve the improvements of +17.3\% and +1.7\% $\mathrm{PQ}^{\text{scene}}$ on ScanNet v2 and Replica.
When using the more advanced open-vocabulary semantic segmentation method, FC-CLIP\cite{yu2023fcclip},
Cues3D further improves $\mathrm{PQ}^{\text{scene}}$ by 6.7\% and 3.5\% on ScanNet v2 and Replica, respectively,
while mIoU increased by 18.2\% and 14.6\% on the two datasets respectively.
Note that Cues3D did not follow the PVLFF setting \cite{pvlff}.
Thus, for an undoubtedly fair comparison,
we conduct an additional evaluation in the same experimental setting as PVLFF \cite{pvlff},
which can be found in \ref{appendix:compare_pvlff}.

\begin{table}[t]
    \centering
        \begin{subtable}{0.53\textwidth}
        \centering
        
        \resizebox{1\linewidth}{!}{
        \begin{tabular}{lccccc}
            \toprule
            \multirow{2.5}{*}{\textbf{Method}} & \multirow{2.5}{*}{\begin{tabular}{c}\textbf{Image}\\\textbf{Feature}\end{tabular}} & \multicolumn{2}{c}{\textbf{ScanNet v2}} & \multicolumn{2}{c}{\textbf{Replica}} \\
            \cmidrule(r){3-4} \cmidrule(r){5-6}
            & & \textbf{mIoU} & \textbf{mAcc} & \textbf{mIoU} & \textbf{mAcc} \\
            \midrule
            LERF \cite{kerr2023lerf} & CLIP & 25.6 & 46.1 & 17.1 & 35.0 \\
            LangSplat \cite{qin2023langsplat} & CLIP & 20.5 & 33.7 & 14.1 & 23.2 \\
            PVLFF \cite{pvlff} & LSeg & \underline{39.7} & 48.5 & 26.2 & 35.0 \\
            \midrule
            \textbf{Cues3D} & CLIP & 38.5 & \underline{52.2} & \underline{27.3} & \underline{35.5} \\
            \textbf{Cues3D} & FC-CLIP & \textbf{56.7} & \textbf{66.6} & \textbf{41.9} & \textbf{53.1} \\
            \bottomrule
        \end{tabular}}
        \caption{Comparing with the image-based methods on 2D metrics.}
        \label{table:semantic_nvs}
    \end{subtable}
    \hfill
    \begin{subtable}{0.46\textwidth}
        \centering
        
        \resizebox{1\linewidth}{!}{
        \begin{tabular}{lccccc}
            \toprule
            \multirow{2.5}{*}{\textbf{Method}} & \multirow{2.5}{*}{\begin{tabular}{c}\textbf{Image}\\\textbf{Feature}\end{tabular}} & \multicolumn{2}{c}{\textbf{ScanNet v2}} \\
            \cmidrule(r){3-4}
            & & \textbf{mIoU} & \textbf{mAcc} \\
            \midrule
            OpenScene (Fusion) \cite{peng2023openscene} & OpenSeg & 24.4 & 39.3 \\
            OpenScene (Distill) \cite{peng2023openscene} & OpenSeg & 33.7 & 52.1 \\
            OpenScene (Ensemble) \cite{peng2023openscene} & OpenSeg & 34.0 & \underline{54.5} \\
            \midrule
            \textbf{Cues3D} & CLIP & 24.1 & 40.2 \\
            \textbf{Cues3D} & OpenSeg & \underline{36.3} & \underline{54.5} \\
            \textbf{Cues3D} & FC-CLIP & \textbf{43.7} & \textbf{68.7} \\
            \bottomrule
        \end{tabular}}
        \caption{Comparing with the 2D-3D merging based methods on 3D metrics.}
        \label{table:semantic_pc}
    \end{subtable}
    \caption{\textbf{Results of open-vocabulary 3D semantic segmentation}.
    Best results are in bold, second best are underlined.}
\end{table}


\begin{figure}[h]
\centering
\includegraphics[width=1\textwidth]{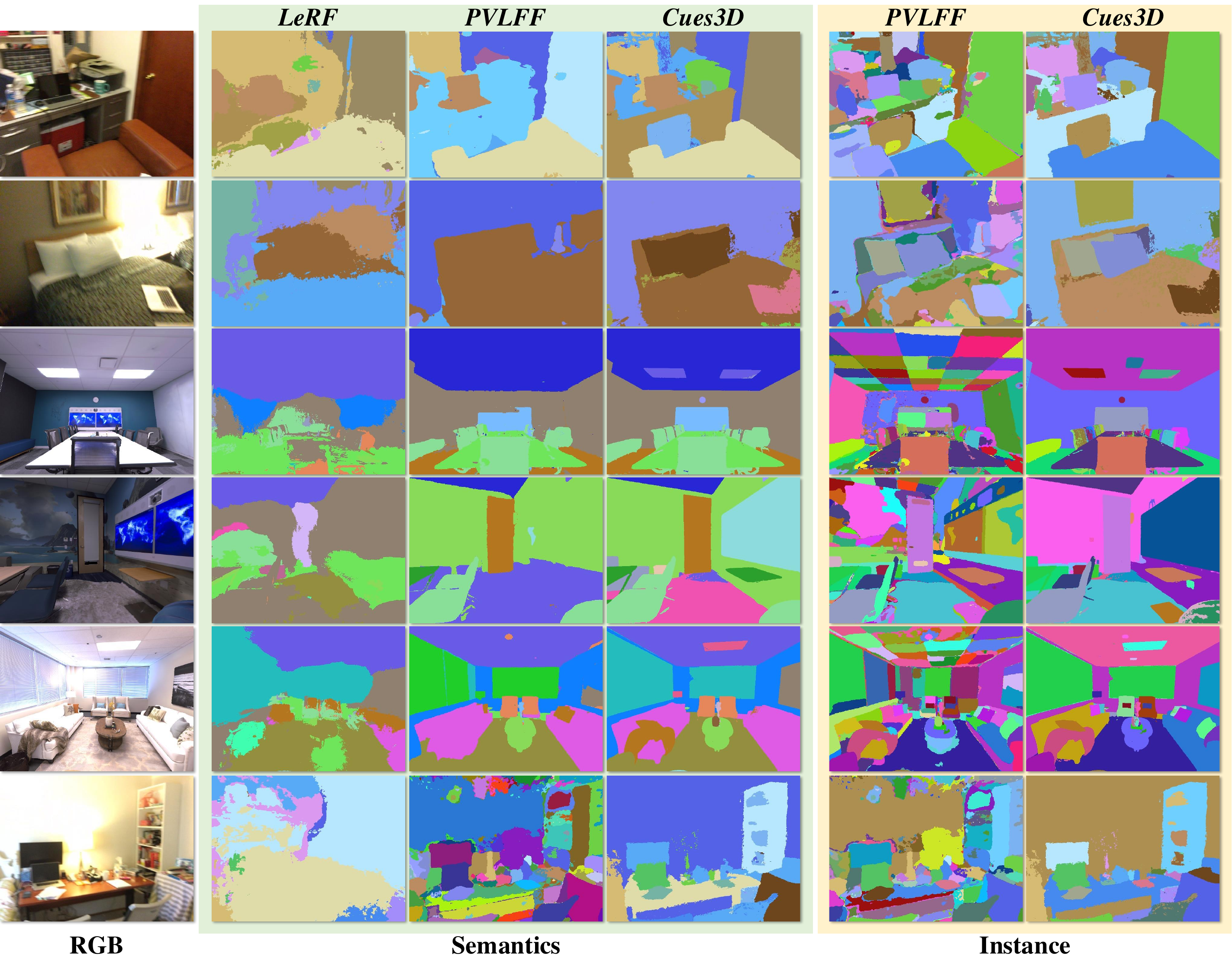}
\caption{\textbf{2D qualitative comparison of NVS.}
We present instance and semantic predictions of 2D image-based methods.}
\label{fig:quali_2d}
\end{figure}

\paragraph{Results of 3D Semantic Segmentation}
Table \ref{table:semantic_nvs} compares 2D image-based methods on the open-vocabulary 3D semantic segmentation.
We exclude three methods for the following reasons:
LEGaussian \cite{shi2023language} cannot achieve usable features via official quantification;
3D-OVS \cite{liu2023weakly} is replaced by LangSplat \cite{qin2023langsplat} with better performance;
OpenNeRF \cite{engelmann2024opennerf} does not have officially provided inference code..
Compared to LERF\cite{kerr2023lerf} and LangSplat \cite{qin2023langsplat} that also use CLIP \cite{clip},
Cues3D achieves a substantial lead of +12.9\% mIoU on ScanNet v2 and +10.2\% mIoU on Replica.
Compared to PVLFF, which uses LSeg, Cues3D achieves comparable performance on ScanNet using only CLIP and demonstrates a notable improvement of +1.1\% mIoU on Replica.
Table \ref{table:semantic_pc} presents a comparison with the 2D-3D merging based method,
OpenScene\cite{peng2023openscene}.
Under the condition of using the same image feature, we achieve a +2.3\% mIoU improvement on ScanNet v2 without requiring any 3D information. 
When using FC-CLIP,
Cues3D outperforms OpenScene by 11.7\% mIoU and 14.2\% mAcc.

\paragraph{Qualitative Results}
Fig. \ref{fig:quali_2d} compares 2D image-based methods of instance segmentation.
PVLFF \cite{pvlff} suffers from a lot of trivial instance areas caused by insufficient discriminability of features, while Cues3D holds high integrality of instances.
In terms of semantic segmentation,
LeRF \cite{kerr2023lerf} suffers from the inaccurate boundary of semantic areas due to lacking instance perception,
and PVLFF mistakenly merges several objects together.
Cues3D correctly segments 2D objects due to correct instances.

\begin{figure}[t]
\centering
\includegraphics[width=1\textwidth]{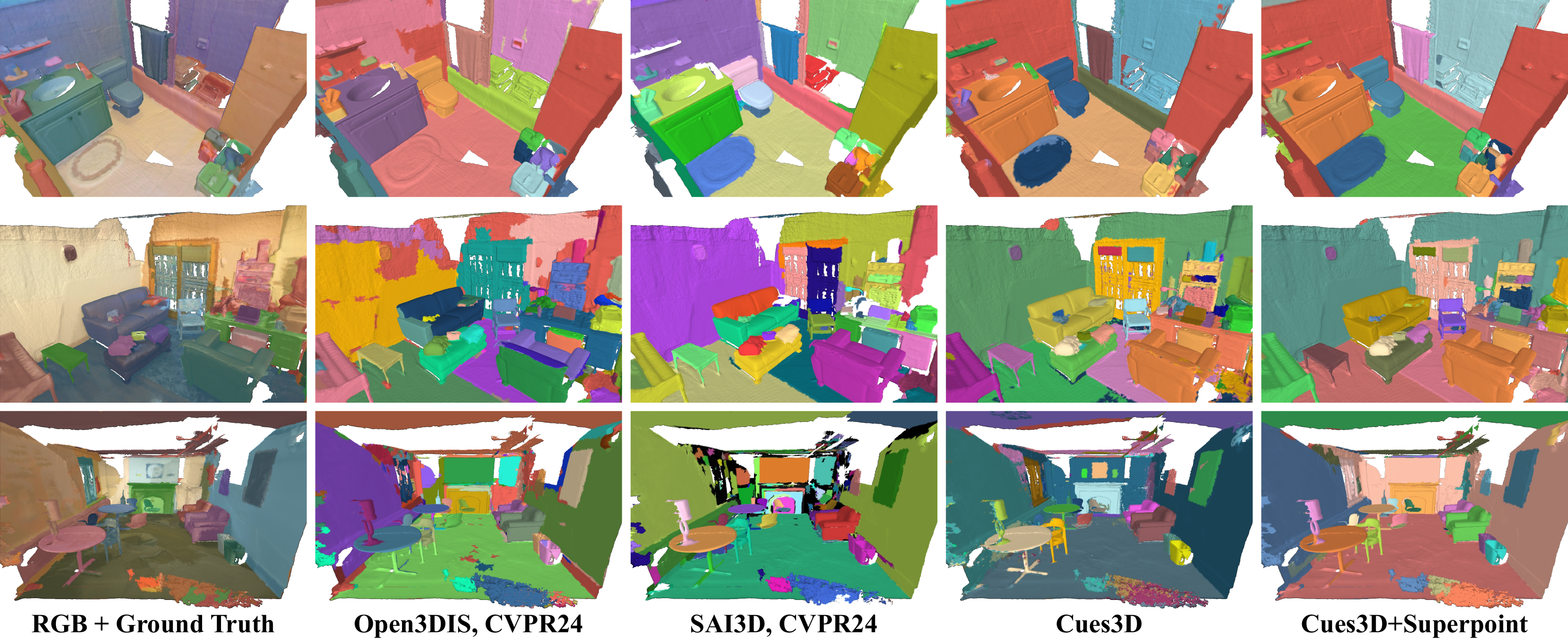}
\caption{\textbf{3D qualitative comparison of zero-shot 3D instance segmentation} (aligned to ground truth point cloud).
Different color indicates distinct instances.}
\label{fig:quali_3d}
\end{figure}

For 3D instance segmentation presented in Fig. \ref{fig:quali_3d},
Cues3D has a better instance integrity for walls than Open3DIS \cite{nguyen2023open3dis} and SAI3D \cite{yin2024sai3d} but only using 2D images.
Benefiting from FC-CLIP \cite{yu2023fcclip},
Cues3D has a better understanding of entire instances,
such as the over-segmented sofas and chairs by Open3DIS and SAI3D.
Note that,
Cues3D may obtain wrong instance IDs at the boundary of objects,
which is due to the misalignment between RGB and Depth data used by 3D visualization.

\begin{table}[t]
\scriptsize
\centering
\resizebox{0.7\linewidth}{!}{
\begin{tabular}{l ccc}
\toprule
\textbf{Setting}      & \textbf{AP} & \textbf{AP50} & \textbf{AP25} \\ 
\midrule
w/o instance disambiguation & 14.4 & 24.3 & 51.1 \\ 
w instance disambiguation       & \textbf{34.9} & \textbf{61.2} & \textbf{76.9} \\ 
\bottomrule
\end{tabular}}
\caption{\textbf{Ablation on instance disambiguation}.
We present the 3D metrics on ScanNet200.}
\label{table:ablation}
\end{table}

\begin{table}[t]
\scriptsize
\centering
\begin{tabular}{lcc|cc}
        \toprule
        \multirow{2}{*}{\textbf{Scene Name}} & \multicolumn{2}{c|}{\textbf{ARI}} & \multicolumn{2}{c}{\textbf{NMI}} \\
        \cmidrule(lr){2-3} \cmidrule(lr){4-5}
        & \textbf{Before} & \textbf{After}  & \textbf{Before} & \textbf{After}  \\
        \midrule
        scene0050\_02 & 39.7 & 42.7 {\scriptsize\textbf{\green{$\uparrow$3.0}}} & 66.8 & 76.8 {\scriptsize\textbf{\green{$\uparrow$10.1}}}\\
        scene0144\_01 & 41.9 & 54.7 {\scriptsize\textbf{\green{$\uparrow$12.8}}} & 66.9 & 81.1 {\scriptsize\textbf{\green{$\uparrow$14.1}}} \\
        scene0221\_01 & 30.7 & 42.0 {\scriptsize\textbf{\green{$\uparrow$11.4}}} & 57.5 & 75.1 {\scriptsize\textbf{\green{$\uparrow$17.5}}} \\
        scene0300\_01 & 41.7 & 51.5 {\scriptsize\textbf{\green{$\uparrow$9.9}}} & 61.1 & 74.4 {\scriptsize\textbf{\green{$\uparrow$13.3}}} \\
        scene0354\_00 & 52.2 & 59.1 {\scriptsize\textbf{\green{$\uparrow$6.9}}} & 64.9 & 73.2 {\scriptsize\textbf{\green{$\uparrow$8.3}}}\\
        scene0389\_00 & 46.2 & 57.3 {\scriptsize\textbf{\green{$\uparrow$11.1}}} & 68.0 & 81.1 {\scriptsize\textbf{\green{$\uparrow$13.1}}} \\
        scene0423\_02 & 95.9 & 95.9 {\scriptsize\textbf{\gray{$\uparrow$0.0}}} & 96.3 & 96.3 {\scriptsize\textbf{\gray{$\uparrow$0.0}}} \\
        scene0427\_00 & 47.4 & 54.5 {\scriptsize\textbf{\green{$\uparrow$7.1}}} & 63.6 & 73.0 {\scriptsize\textbf{\green{$\uparrow$9.4}}} \\
        scene0494\_00 & 50.2 & 51.0 {\scriptsize\textbf{\green{$\uparrow$0.8}}} & 63.9 & 66.9 {\scriptsize\textbf{\green{$\uparrow$3.0}}} \\
        scene0616\_00 & 41.7 & 52.3 {\scriptsize\textbf{\green{$\uparrow$10.6}}}& 60.7 & 74.6 {\scriptsize\textbf{\green{$\uparrow$13.9}}}\\
        scene0645\_02 & 42.3 & 52.7 {\scriptsize\textbf{\green{$\uparrow$10.4}}} & 64.5 & 78.2 {\scriptsize\textbf{\green{$\uparrow$13.7}}} \\
        scene0693\_00 & 43.9 & 51.6 {\scriptsize\textbf{\green{$\uparrow$7.6}}} & 67.1 & 76.0 {\scriptsize\textbf{\green{$\uparrow$8.9}}}\\
        \midrule
        Average & 47.8 & 55.4 {\scriptsize\textbf{\green{$\uparrow$7.6}}} & 37.2 & 77.2 {\scriptsize\textbf{\green{$\uparrow$10.4}}} \\
        \bottomrule
\end{tabular}
\caption{\xf{\textbf{Comparison of ARI and NMI results before and after disambiguation}}.}
\label{table:ari_nmi}
\end{table}

\subsection{Detailed Analysis}
\paragraph{Ablation Study on Instance Disambiguation}
Table \ref{table:ablation} shows the ScanNet200 results with and without instance disambiguation.
If the instance disambiguation is removed,
it does not perform well in 3D instance segmentation tasks.
\xf{The reason is that different objects may mistakenly be assigned the same ID without the $2^{\text{th}}$ stage.}

\xf{To further demonstrate the importance of the second stage,
we use two metrics for comparing results before and after disambiguation on ScanNet200.
Referring from clustering evaluation,
Adjusted Rand Index (ARI) and Normalized Mutual Information (NMI) are used to evaluate whether the division of the 3D mask group is the same as the real one.
As shown in Table \ref{table:ari_nmi},
after the disambiguation stage, the accuracy of 3D ID assignment in each scene improved to varying degrees.
The ARI metric increased by an average of 7.6\%, indicating a significant improvement in correctly grouped 3D masks. The NMI metric rose by 10.4\%,
showing that the processed 3D mask groups align more closely with the ground truth IDs.}

\paragraph{Ablation Study on Matching Criteria}
In Sec. \ref{sec:disambiguation},
$\tau_\mathbf{d}$ and $\tau_\mathbf{n}$ are taken as the parameters for matching.
Table \ref{table:ablation_instance} and \ref{table:ablation_point_num} show the performance using different settings of $\tau_\mathbf{d}$ and $\tau_\mathbf{n}$.
When $\tau_\mathbf{d}$ is 0.05 or 0.075,
Cues3D achieves the best performance.
When $\tau_\mathbf{n}$ is 50 or 75, the performance is the best.
In summary, when $\tau_\mathbf{d}$ is \xf{0.075} and $\tau_\mathbf{n}$ is 50, the model has the best overall performance.

\begin{table}[t]
\scriptsize
    \centering
    \begin{subtable}{0.48\textwidth}
        \centering
        \begin{tabular}{p{0.18\textwidth}p{0.18\textwidth}p{0.18\textwidth}p{0.18\textwidth}}
            \toprule
            $\tau_\mathbf{d}$ & \textbf{AP} & \textbf{AP50} & \textbf{AP25} \\
            \midrule
            0.025 & 32.0 & 55.8 & 71.2 \\
            0.05 & \textbf{35.5} & \underline{60.7} & \underline{74.2} \\
            0.075 & \underline{34.9} & \textbf{61.2} & \textbf{76.9} \\
            0.1 & 33.7 & 57.6 & 73.4 \\
            \bottomrule
        \end{tabular}
        \caption{Ablation on $\tau_\mathbf{d}$.}
        \label{table:ablation_instance}
    \end{subtable}
    \hfill
    \begin{subtable}{0.48\textwidth}
        \centering
        \begin{tabular}{p{0.18\textwidth}p{0.18\textwidth}p{0.18\textwidth}p{0.18\textwidth}}
            \toprule
            $\tau_\mathbf{n}$ & \textbf{AP} & \textbf{AP50} & \textbf{AP25} \\
            \midrule
            0 & 31.6 & 52.9 & 69.8 \\
            25 & 33.5 & 57.8 & 73.0 \\
            50 & \underline{34.9} & \textbf{61.2} & \textbf{76.9} \\
            75 & \textbf{35.5} & \underline{60.8} & \underline{74.2} \\
            \bottomrule
        \end{tabular}
        \caption{Ablation on $\tau_\mathbf{n}$.}
        \label{table:ablation_point_num}
    \end{subtable}
    \caption{\textbf{Ablation on matching criteria.}
    We present the performance on ScanNet200.}
\end{table}

\paragraph{Ablation Study on Mask Order}
In the disambiguation phase of Cues3D,
the 3D masks in a group are sorted according to the image sequence.
To investigate whether the mask order affects the results,
we input the images in a random order for comparison.
Table \ref{table:merge_order} shows the corresponding performance.
Observably, the APs of using the two orders are similar.
The reason is that,
all 3D masks in $\mathbf{G}_u$ would be combined into a unified point cloud eventually,
and whether two 3D masks belong to the same object is determined solely by the number of overlapping points, regardless of their input order.
Therefore, the mask order does not significantly impact our second phase.

\begin{table}[t]
\scriptsize
\centering
\begin{tabular}{l ccc}
\toprule
\textbf{Mask order in the group $\mathbf{G}_u$}      & \textbf{AP} & \textbf{AP50} & \textbf{AP25} \\ 
\midrule
Timestamp order         & 34.9 & \textbf{61.2} & \textbf{76.9} \\ 
Randomly shuffled order & \textbf{35.4} & 59.7 & 73.7 \\ 
\bottomrule
\end{tabular}
\caption{\textbf{Impact of merging order}.
We present the performance of different orders on ScanNet200.}
\label{table:merge_order}
\end{table}

\paragraph{Impact of Training Iteration Variation}
Since NeRFacto fully converges in approximately 30,000 iterations,
we split such 30,000 iterations into two parts for the first and third phases.
Table \ref{table:iteration_division} shows the impact on performance when using different iteration splits.
When there are only 10,000 iterations of the first phase, Cues3D obtain only 31.1\% AP on ScanNet200.
As the iterations for the first phase increase from 10,000 to 20,000, the AP increases in succession until 34.9\%
Additionally, if there are only 5,000 iterations of the second phase,
the model suffers from insufficient optimization and fails to achieve ideal results.
When using a 20,000:10,000 split,
Cues3D achieves a good balance between assigning consistent instance IDs across multiple views and performing sufficient optimization.
\paragraph{Time Consumption}
Table \ref{table:time_consumption} shows the time cost of each phase.
All the time is tested in the scenes of Replica dataset.
Each contains 240 training images.
The 2$^\text{nd}$ phase takes only 13\% of the overall training time and does not significantly extend the total time required for the entire process.
3D Mask Extraction takes a longer time than disambiguation because it requires prediction on all training images.
Therefore, if a faster NVS is employed,
3D Mask Extraction will be greatly accelerated.

\begin{table}[t]
    \centering
        \begin{subtable}{0.465\textwidth}
        \centering
        
        \resizebox{1\linewidth}{!}{
        \begin{tabular}{lccc}
            \toprule
            \textbf{Iteration split} & \textbf{AP} & \textbf{AP50} & \textbf{AP25} \\
            \midrule
            10,000+20,000 & 31.1 & 55.1 & 67.8 \\
            15,000+15,000 & \underline{32.0} & \underline{56.1} & \underline{74.0} \\
            \textbf{20,000+10,000} & \textbf{34.9} & \textbf{61.2} & \textbf{76.9} \\
            25,000+5,000 & 31.0 & 53.1 & 68.1 \\
            \bottomrule
            \end{tabular}
        }
        \caption{Results with iteration settings on ScanNet200. }
        \label{table:iteration_division}
    \end{subtable}
    \hfill
    \begin{subtable}{0.525\textwidth}
        \centering
        \resizebox{1\linewidth}{!}{
            \begin{tabular}{ll}
            \toprule
            \textbf{Training phases} & \textbf{time} \\
            \midrule
            1$^\text{st}$: Instance Field Initialization & 35.5 minutes \\
            2$^\text{nd}$: 3D Masks Extraction & 4.9 minutes \\
            2$^\text{nd}$: Instance Disambiguation & 2.7 minutes \\
            3$^\text{rd}$: Instance Field Fine-Tuning & 13.3 minutes \\
            \bottomrule
            \end{tabular}
        }
        \caption{The detailed time consumption of each phase.}
        \label{table:time_consumption}
    \end{subtable}
    \caption{\textbf{Detailed analysis of Cues3D.}
    Training iteration impact and training time statistics.}
\end{table}

\begin{figure}[t]
\centering
\includegraphics[width=1\textwidth]{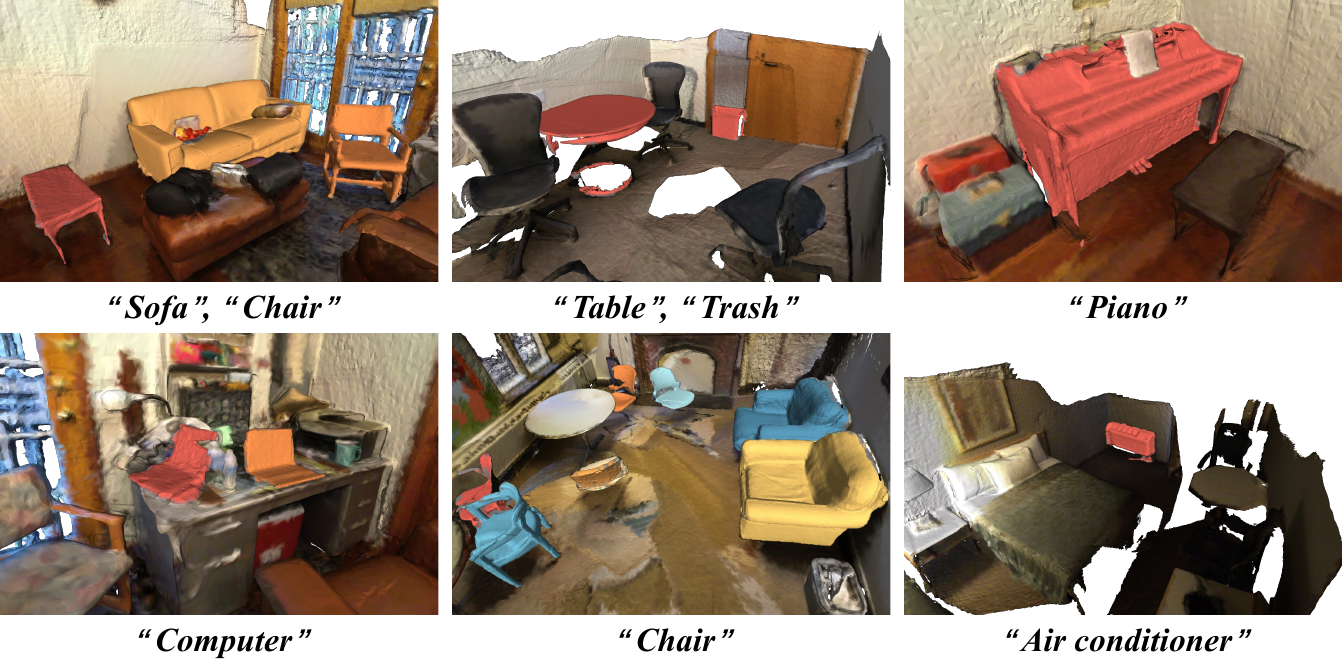}
\caption{\textbf{Visualization of object query results.}
Different objects in the same scene are visualized in different colors.}
\label{fig:query}
\end{figure}

\paragraph{Open-Vocabulary 3D Object Query}
In addition to performing open-vocabulary 3D instance segmentation and semantic segmentation,
we also query 3D objects in a scene using input prompts.
Note that,
we match the queries with the multi-view masks of each object,
which allows us to find the objects with the highest similarities.
Fig. \ref{fig:query} shows the visualization results of querying objects in an open-vocabulary manner.
Cues3D can find all the objects to be queried in the scene and obtain a clear boundary of objects.

\begin{figure}[t]
\centering
\includegraphics[width=1\textwidth]{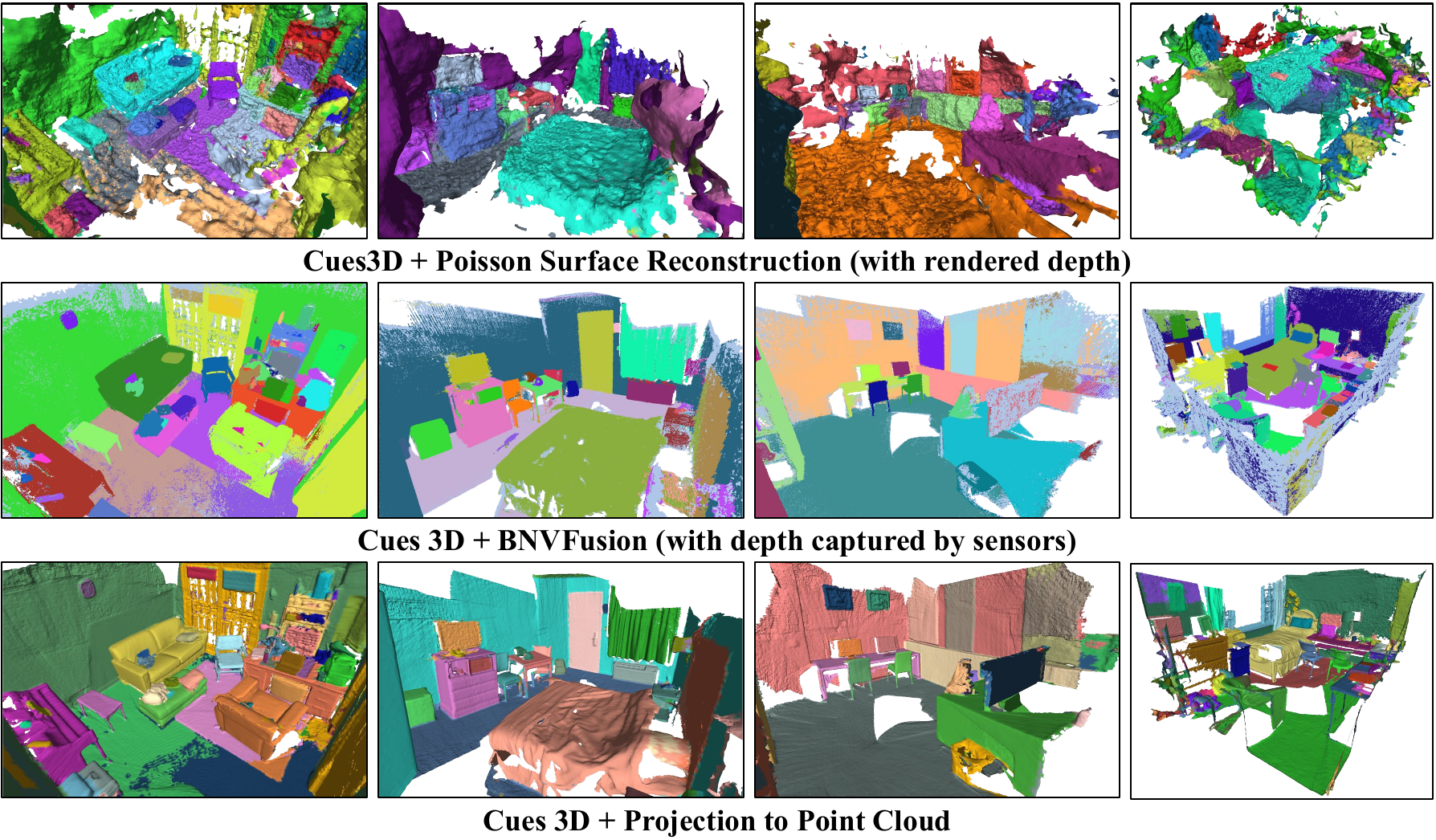}
\caption{\textbf{3D reconstruction visualizations.}
The consistent instances of the 3D Stable are constructed using Poisson reconstruction \cite{kazhdan2006poisson}, BNV-Fusion \cite{li2022bnv} and projection to the point cloud respectively.}
\label{fig:reconstruction}
\end{figure}

\paragraph{3D Visualizations using Different Reconstruction Methods}
Our method is based on NVS that does not use depth captured by sensors,
which poses a significant challenge to reconstruction quality and comparison.
Therefore,
when evaluating against methods based on 2D-3D merging,
we project the prediction of Cues3D onto the reconstructed point cloud for a feasible comparison.
Here, we also visualize the 3D representation reconstructed by commonly used Poisson surface reconstruction \cite{kazhdan2006poisson}.
In the first row of Fig. \ref{fig:reconstruction},
due to the lack of supervision from depth information,
it can only generate a rough 3D representation but still has reasonable instances.
This explains why we need to first initialize the instance field and then proceed with point cloud matching,
as we cannot directly use a low-quality 3D representation for point cloud matching.
We also use BNV-Fusion \cite{li2022bnv} to reconstruct 3D instances,
which uses metric depth data from depth sensors, as shown in the second row.
Such results show that Cues3D equipped with a depth camera can obtain a good 3D instance segmentation result.
In the last row,
Cues3D can achieve similar results to 2D-3D based methods after being projected onto the ground truth point cloud.

\section{Conclusion}

This paper proposes Cues3D,
a compact and efficient zero-shot 3D instance segmentation algorithm to improve open-vocabulary 3D panoptic segmentation.
Our core idea is to use the global view of NeRF's 3D implicit field to distinguish objects.
Specifically, Cues3D contains a three-stage training framework. After the first initialization phase,
the second phase uses the 3D knowledge learned in the first phase to correct the ID of the instance to achieve the uniqueness of the ID.
In the third stage, corrected instances are directly used to train a stable instance field.
Extensive experiments on the three tasks of instance, panoptic, and semantic segmentation verify the effectiveness of Cues3D.


\paragraph{Limitations and Furture Works}
\xf{
Our limitations are mainly twofold.
Firstly, since disambiguation relies on NeRF-generated 3D representations,
poor 3D reconstruction quality may disrupt the disambiguation process,
although this issue occurs in only a few scenes.
This may limit the applicability of our method on poor sensors,
fast-motion scenes, and poor reconstruction basis.
Secondly, our current approach does not support hierarchical 3D instance segmentation, preventing flexible 3D semantic predictions. This may limit the application of the proposed algorithm in future robot-environment interactions. 

To address the first issue,
the performance of Cues3D can be further improved with the emergence of higher-performing NVS methods.
For the second issue, we suggest to build an online 3D scene understanding system, which continuously updates 3D instance representations of scenes. Due to the complexity of the real world, building an exhaustive hierarchical 3D instance representation may be computationally expensive. An online 3D scene understanding system can only perform local 3D understanding and updates on the 3D scene seen by the AI agent, which is  efficient.


According to the characteristics of Cues3D,
Cues3D is particularly well-suited for applications requiring high-fidelity 3D scene reconstruction under stable camera trajectories,
such as indoor navigation systems, domestic service robotics, and interior decoration.
By leveraging its precise 3D mask, Cues3D can enhance the spatial understanding of robots navigating cluttered indoor spaces or improve the accuracy of virtual room planning in interior design applications.

Future research could explore extending Cues3D to handle dynamic environments where objects and agents move unpredictably.
Additionally, integrating multi-modal data, such as RGB and LiDAR,
could further improve robustness in challenging lighting and fast-moving conditions.
Another promising direction is optimizing the model’s efficiency for real-time deployment on edge devices,
enabling broader accessibility in resource-constrained scenarios.
}

\bibliographystyle{plain}
\bibliography{main_v3_major}

\begin{thebibliography}{10}

\bibitem{barron2021mip}
Jonathan~T Barron, Ben Mildenhall, Matthew Tancik, Peter Hedman, Ricardo Martin-Brualla, and Pratul~P Srinivasan.
\newblock Mip-nerf: A multiscale representation for anti-aliasing neural radiance fields.
\newblock In {\em Proceedings of the IEEE/CVF International Conference on Computer Vision (ICCV)}, 2021.

\bibitem{barron2022mip}
Jonathan~T Barron, Ben Mildenhall, Dor Verbin, Pratul~P Srinivasan, and Peter Hedman.
\newblock Mip-nerf 360: Unbounded anti-aliased neural radiance fields.
\newblock In {\em Proceedings of the IEEE/CVF Conference on Computer Vision and Pattern Recognition (CVPR)}, 2022.

\bibitem{bhalgat2023contrastive}
Yash~Sanjay Bhalgat, Iro Laina, Joao~F. Henriques, Andrea Vedaldi, and Andrew Zisserman.
\newblock Contrastive lift: 3d object instance segmentation by slow-fast contrastive fusion.
\newblock In {\em Conference on Neural Information Processing Systems (NeurIPS)}, 2023.

\bibitem{pvlff}
Haoran Chen, Kenneth Blomqvist, Francesco Milano, and Roland Siegwart.
\newblock Panoptic vision-language feature fields.
\newblock {\em IEEE Robotics and Automation Letters}, 9(3):2144 -- 2151, 2024.

\bibitem{chen2024robustsam}
Wei-Ting Chen, Yu-Jiet Vong, Sy-Yen Kuo, Sizhou Ma, and Jian Wang.
\newblock Robustsam: Segment anything robustly on degraded images.
\newblock In {\em IEEE/CVF Conference on Computer Vision and Pattern Recognition (CVPR)}, 2024.

\bibitem{dai2017scannet}
Angela Dai, Angel~X Chang, Manolis Savva, Maciej Halber, Thomas Funkhouser, and Matthias Nie{\ss}ner.
\newblock Scannet: Richly-annotated 3d reconstructions of indoor scenes.
\newblock In {\em Proceedings of the IEEE/CVF Conference on Computer Vision and Pattern Recognition (CVPR)}, 2017.

\bibitem{zegformer}
Jian Ding, Nan Xue, Gui-Song Xia, and Dengxin Dai.
\newblock Decoupling zero-shot semantic segmentation.
\newblock In {\em IEEE/CVF Conference on Computer Vision and Pattern Recognition (CVPR)}, 2022.

\bibitem{engelmann2024opennerf}
Francis Engelmann, Fabian Manhardt, Michael Niemeyer, Keisuke Tateno, Marc Pollefeys, and Federico Tombari.
\newblock {OpenNeRF: Open Set 3D Neural Scene Segmentation with Pixel-Wise Features and Rendered Novel Views}.
\newblock In {\em International Conference on Learning Representations (ICLR)}, 2024.

\bibitem{ghiasi2021open}
Golnaz Ghiasi, Xiuye Gu, Yin Cui, and Tsung-Yi Lin.
\newblock Scaling open-vocabulary image segmentation with image-level labels.
\newblock In {\em European Conference on Computer Vision (ECCV)}, 2022.

\bibitem{10160969}
Chenguang Huang, Oier Mees, Andy Zeng, and Wolfram Burgard.
\newblock Visual language maps for robot navigation.
\newblock In {\em IEEE International Conference on Robotics and Automation (ICRA)}, 2023.

\bibitem{jiang2023pointgs}
Chenru Jiang, Kaizhu Huang, Junwei Wu, Xinheng Wang, Jimin Xiao, and Amir Hussain.
\newblock Pointgs: Bridging and fusing geometric and semantic space for 3d point cloud analysis.
\newblock {\em Information Fusion}, 91:316--326, 2023.

\bibitem{kajiya1984ray}
James~T. Kajiya and Brian~P Von~Herzen.
\newblock Ray tracing volume densities.
\newblock {\em SIGGRAPH Comput. Graph.}, 18(3):165–174, Jan 1984.

\bibitem{kazhdan2006poisson}
Michael Kazhdan, Matthew Bolitho, and Hugues Hoppe.
\newblock Poisson surface reconstruction.
\newblock In {\em Eurographics symposium on Geometry processing}, 2006.

\bibitem{kerbl20233d}
Bernhard Kerbl, Georgios Kopanas, Thomas Leimk{\"u}hler, and George Drettakis.
\newblock 3d gaussian splatting for real-time radiance field rendering.
\newblock {\em ACM Transactions on Graphics}, 42(4):1--14, 2023.

\bibitem{kerr2023lerf}
Justin Kerr, Chung~Min Kim, Ken Goldberg, Angjoo Kanazawa, and Matthew Tancik.
\newblock Lerf: Language embedded radiance fields.
\newblock In {\em Proceedings of the IEEE/CVF International Conference on Computer Vision (ICCV)}, 2023.

\bibitem{sam}
Alexander Kirillov, Eric Mintun, Nikhila Ravi, Hanzi Mao, Chloe Rolland, Laura Gustafson, Tete Xiao, Spencer Whitehead, Alexander~C Berg, Wan-Yen Lo, et~al.
\newblock Segment anything.
\newblock In {\em Proceedings of the IEEE/CVF International Conference on Computer Vision (ICCV)}, 2023.

\bibitem{kuhn1955hungarian}
Harold~W Kuhn.
\newblock The hungarian method for the assignment problem.
\newblock {\em Naval research logistics quarterly}, 2(1-2):83--97, 1955.

\bibitem{9879705}
Xin Lai, Jianhui Liu, Li~Jiang, Liwei Wang, Hengshuang Zhao, Shu Liu, Xiaojuan Qi, and Jiaya Jia.
\newblock Stratified transformer for 3d point cloud segmentation.
\newblock In {\em IEEE/CVF Conference on Computer Vision and Pattern Recognition (CVPR)}, 2022.

\bibitem{lseg}
Boyi Li, Kilian~Q Weinberger, Serge Belongie, Vladlen Koltun, and Rene Ranftl.
\newblock Language-driven semantic segmentation.
\newblock In {\em International Conference on Learning Representations (ICLR)}, 2022.

\bibitem{li2022bnv}
Kejie Li, Yansong Tang, Victor~Adrian Prisacariu, and Philip~HS Torr.
\newblock Bnv-fusion: Dense 3d reconstruction using bi-level neural volume fusion.
\newblock In {\em IEEE/CVF Conference on Computer Vision and Pattern Recognition (CVPR)}, 2022.

\bibitem{openseg}
Feng Liang, Bichen Wu, Xiaoliang Dai, Kunpeng Li, Yinan Zhao, Hang Zhang, Peizhao Zhang, Peter Vajda, and Diana Marculescu.
\newblock Open-vocabulary semantic segmentation with mask-adapted clip.
\newblock In {\em IEEE/CVF Conference on Computer Vision and Pattern Recognition (CVPR)}, 2023.

\bibitem{liang2023unknown}
Wenteng Liang, Feng Xue, Yihao Liu, Guofeng Zhong, and Anlong Ming.
\newblock Unknown sniffer for object detection: Don't turn a blind eye to unknown objects.
\newblock In {\em IEEE/CVF Conference on Computer Vision and Pattern Recognition}, 2023.

\bibitem{liu2023weakly}
Kunhao Liu, Fangneng Zhan, Jiahui Zhang, Muyu Xu, Yingchen Yu, Abdulmotaleb~El Saddik, Christian Theobalt, Eric Xing, and Shijian Lu.
\newblock Weakly supervised 3d open-vocabulary segmentation.
\newblock In {\em Conference on Neural Information Processing Systems (NeurIPS)}, 2023.

\bibitem{mildenhall2021nerf}
Ben Mildenhall, Pratul~P Srinivasan, Matthew Tancik, Jonathan~T Barron, Ravi Ramamoorthi, and Ren Ng.
\newblock Nerf: Representing scenes as neural radiance fields for view synthesis.
\newblock {\em Communications of the ACM}, 65(1):99--106, 2021.

\bibitem{mueller2022instant}
Thomas M\"uller, Alex Evans, Christoph Schied, and Alexander Keller.
\newblock Instant neural graphics primitives with a multiresolution hash encoding.
\newblock {\em ACM Transactions on Graphics (ToG)}, 41(4):102:1--102:15, July 2022.

\bibitem{nguyen2023open3dis}
Phuc Nguyen, Tuan~Duc Ngo, Evangelos Kalogerakis, Chuang Gan, Anh Tran, Cuong Pham, and Khoi Nguyen.
\newblock Open3dis: Open-vocabulary 3d instance segmentation with 2d mask guidance.
\newblock In {\em Proceedings of the IEEE/CVF Conference on Computer Vision and Pattern Recognition (CVPR)}, 2024.

\bibitem{peng2023openscene}
Songyou Peng, Kyle Genova, Chiyu Jiang, Andrea Tagliasacchi, Marc Pollefeys, Thomas Funkhouser, et~al.
\newblock Openscene: 3d scene understanding with open vocabularies.
\newblock In {\em Proceedings of the IEEE/CVF Conference on Computer Vision and Pattern Recognition (CVPR)}, 2023.

\bibitem{qi2017pointnet}
Charles~R Qi, Hao Su, Kaichun Mo, and Leonidas~J Guibas.
\newblock Pointnet: Deep learning on point sets for 3d classification and segmentation.
\newblock In {\em IEEE/CVF Conference on Computer Vision and Pattern Recognition (CVPR)}, 2017.

\bibitem{cropformer}
Lu~Qi, Jason Kuen, Tiancheng Shen, Jiuxiang Gu, Weidong Guo, Jiaya Jia, Zhe Lin, and Ming-Hsuan Yang.
\newblock High-quality entity segmentation.
\newblock In {\em Proceedings of the IEEE/CVF International Conference on Computer Vision (ICCV)}, 2023.

\bibitem{qin2023langsplat}
Minghan Qin, Wanhua Li, Jiawei Zhou, Haoqian Wang, and Hanspeter Pfister.
\newblock Langsplat: 3d language gaussian splatting.
\newblock In {\em Proceedings of the IEEE/CVF Conference on Computer Vision and Pattern Recognition (CVPR)}, 2024.

\bibitem{clip}
Alec Radford, Jong~Wook Kim, Chris Hallacy, Aditya Ramesh, Gabriel Goh, Sandhini Agarwal, Girish Sastry, Amanda Askell, Pamela Mishkin, Jack Clark, et~al.
\newblock Learning transferable visual models from natural language supervision.
\newblock In {\em International conference on machine learning (ICML)}, 2021.

\bibitem{rozenberszki2022language}
David Rozenberszki, Or~Litany, and Angela Dai.
\newblock Language-grounded indoor 3d semantic segmentation in the wild.
\newblock In {\em European Conference on Computer Vision (ECCV)}, 2022.

\bibitem{shi2023language}
Jin-Chuan Shi, Miao Wang, Hao-Bin Duan, and Shao-Hua Guan.
\newblock Language embedded 3d gaussians for open-vocabulary scene understanding.
\newblock In {\em Proceedings of the IEEE/CVF Conference on Computer Vision and Pattern Recognition (CVPR)}, 2024.

\bibitem{2023_CVPR_panopticlifting}
Yawar Siddiqui, Lorenzo Porzi, Samuel~Rota Bul{\'o}, Norman M{\"u}ller, Matthias Nie{\ss}ner, Angela Dai, and Peter Kontschieder.
\newblock Panoptic lifting for 3d scene understanding with neural fields.
\newblock In {\em Proceedings of the IEEE/CVF Conference on Computer Vision and Pattern Recognition (CVPR)}, 2023.

\bibitem{samlightening}
Yanfei Song, Bangzheng Pu, Peng Wang, Hongxu Jiang, Dong Dong, Yongxiang Cao, and Yiqing Shen.
\newblock Sam-lightening: A lightweight segment anything model with dilated flash attention to achieve 30 times acceleration.
\newblock {\em arXiv preprint arXiv:2403.09195}, 2024.

\bibitem{replica19arxiv}
Julian Straub, Thomas Whelan, Lingni Ma, Yufan Chen, Erik Wijmans, Simon Green, Jakob~J Engel, Raul Mur-Artal, Carl Ren, Shobhit Verma, et~al.
\newblock The replica dataset: A digital replica of indoor spaces.
\newblock {\em arXiv preprint arXiv:1906.05797}, 2019.

\bibitem{takmaz2023openmaskd}
Ay{\c{c}}a Takmaz, Elisabetta Fedele, Robert Sumner, Marc Pollefeys, Federico Tombari, and Francis Engelmann.
\newblock Openmask3d: Open-vocabulary 3d instance segmentation.
\newblock In {\em Conference on Neural Information Processing Systems (NeurIPS)}, 2023.

\bibitem{nerfstudio}
Matthew Tancik, Ethan Weber, Evonne Ng, Ruilong Li, Brent Yi, Justin Kerr, Terrance Wang, Alexander Kristoffersen, Jake Austin, Kamyar Salahi, Abhik Ahuja, David McAllister, and Angjoo Kanazawa.
\newblock Nerfstudio: A modular framework for neural radiance field development.
\newblock In {\em ACM SIGGRAPH 2023 Conference Proceedings}, 2023.

\bibitem{iml}
George Tang, Krishna~Murthy Jatavallabhula, and Antonio Torralba.
\newblock Efficient 3d instance mapping and localization with neural fields.
\newblock In {\em International Conference on Robotics and Automation (ICRA)}, 2024.

\bibitem{wang2023dmnerf}
Bing WANG, Lu~Chen, and Bo~Yang.
\newblock {DM}-ne{RF}: 3d scene geometry decomposition and manipulation from 2d images.
\newblock In {\em International Conference on Learning Representations (ICLR)}, 2023.

\bibitem{tiny_vit}
Kan Wu, Jinnian Zhang, Houwen Peng, Mengchen Liu, Bin Xiao, Jianlong Fu, and Lu~Yuan.
\newblock Tinyvit: Fast pretraining distillation for small vision transformers.
\newblock In {\em European conference on computer vision (ECCV)}, 2022.

\bibitem{WU2024102532}
Yuchen Wu, Pengcheng Zhang, Meiying Gu, Jin Zheng, and Xiao Bai.
\newblock Embodied navigation with multi-modal information: A survey from tasks to methodology.
\newblock {\em Information Fusion}, 112:102532, 2024.

\bibitem{10249213}
Yue Wu, Jiaming Liu, Maoguo Gong, Zhixiao Liu, Qiguang Miao, and Wenping Ma.
\newblock Mpct: Multiscale point cloud transformer with a residual network.
\newblock {\em IEEE Transactions on Multimedia}, 26:3505--3516, 2024.

\bibitem{wu2024joint}
Yue Wu, Jiaming Liu, Maoguo Gong, Qiguang Miao, Wenping Ma, and Cai Xu.
\newblock Joint semantic segmentation using representations of lidar point clouds and camera images.
\newblock {\em Information Fusion}, 108:102370, 2024.

\bibitem{10566881}
Yue Wu, Jinlong Sheng, Hangqi Ding, Peiran Gong, Hao Li, Maoguo Gong, Wenping Ma, and Qiguang Miao.
\newblock Evolutionary multitasking descriptor optimization for point cloud registration.
\newblock {\em IEEE Transactions on Evolutionary Computation}, pages 1--1, 2024.

\bibitem{EfficientSAM}
Yunyang Xiong, Bala Varadarajan, Lemeng Wu, Xiaoyu Xiang, Fanyi Xiao, Chenchen Zhu, Xiaoliang Dai, Dilin Wang, Fei Sun, Forrest Iandola, Raghuraman Krishnamoorthi, and Vikas Chandra.
\newblock Efficientsam: Leveraged masked image pretraining for efficient segment anything.
\newblock In {\em IEEE/CVF Conference on Computer Vision and Pattern Recognition (CVPR)}, 2024.

\bibitem{ICRA}
F.~{Xue}, A.~{Ming}, M.~{Zhou}, and Y.~{Zhou}.
\newblock A novel multi-layer framework for tiny obstacle discovery.
\newblock In {\em IEEE International Conference on Robotics and Automation (ICRA)}, 2019.

\bibitem{xue_indoor_2023}
Feng Xue, Yicong Chang, Tianxi Wang, Yu~Zhou, and Anlong Ming.
\newblock Indoor {Obstacle} {Discovery} on {Reflective} {Ground} via {Monocular} {Camera}.
\newblock {\em International Journal of Computer Vision (IJCV)}, 132:987–1007, 2023.

\bibitem{yan2024maskclustering}
Mi~Yan, Jiazhao Zhang, Yan Zhu, and He~Wang.
\newblock Maskclustering: View consensus based mask graph clustering for open-vocabulary 3d instance segmentation.
\newblock In {\em Proceedings of the IEEE/CVF Conference on Computer Vision and Pattern Recognition (CVPR)}, 2024.

\bibitem{yang2019learning}
Bo~Yang, Jianan Wang, Ronald Clark, Qingyong Hu, Sen Wang, Andrew Markham, and Niki Trigoni.
\newblock Learning object bounding boxes for 3d instance segmentation on point clouds.
\newblock In {\em Advances in neural information processing systems (NeurIPS)}, 2019.

\bibitem{gaussian_grouping}
Mingqiao Ye, Martin Danelljan, Fisher Yu, and Lei Ke.
\newblock Gaussian grouping: Segment and edit anything in 3d scenes.
\newblock {\em arXiv preprint arXiv:2312.00732}, 2023.

\bibitem{yeshwanthliu2023scannetpp}
Chandan Yeshwanth, Yueh-Cheng Liu, Matthias Nie{\ss}ner, and Angela Dai.
\newblock Scannet++: A high-fidelity dataset of 3d indoor scenes.
\newblock In {\em Proceedings of the International Conference on Computer Vision (ICCV)}, 2023.

\bibitem{yin2024sai3d}
Yingda Yin, Yuzheng Liu, Yang Xiao, Daniel Cohen-Or, Jingwei Huang, and Baoquan Chen.
\newblock Sai3d: Segment any instance in 3d scenes.
\newblock In {\em Proceedings of the IEEE/CVF Conference on Computer Vision and Pattern Recognition (CVPR)}, 2024.

\bibitem{yu2023fcclip}
Qihang Yu, Ju~He, Xueqing Deng, Xiaohui Shen, and Liang-Chieh Chen.
\newblock Convolutions die hard: Open-vocabulary segmentation with single frozen convolutional clip.
\newblock In {\em Conference on Neural Information Processing Systems (NeurIPS)}, 2023.

\bibitem{10319695}
Yongzhe Yuan, Yue Wu, Xiaolong Fan, Maoguo Gong, Wenping Ma, and Qiguang Miao.
\newblock Egst: Enhanced geometric structure transformer for point cloud registration.
\newblock {\em IEEE Transactions on Visualization and Computer Graphics}, 30(9):6222--6234, 2024.

\bibitem{yang2023sam3d}
Yang Yunhan, Wu~Xiaoyang, He~Tong, Zhao Hengshuang, and Liu Xihui.
\newblock Sam3d: Segment anything in 3d scenes.
\newblock In {\em Proceedings of the IEEE/CVF International Conference on Computer Vision Workshop (ICCV Workshop)}, 2023.

\bibitem{mobilesam}
Chaoning Zhang, Dongshen Han, Yu~Qiao, Jung~Uk Kim, Sung-Ho Bae, Seungkyu Lee, and Choong~Seon Hong.
\newblock Faster segment anything: Towards lightweight sam for mobile applications.
\newblock {\em arXiv preprint arXiv:2306.14289}, 2023.

\bibitem{9710703}
Hengshuang Zhao, Li~Jiang, Jiaya Jia, Philip Torr, and Vladlen Koltun.
\newblock Point transformer.
\newblock In {\em IEEE/CVF International Conference on Computer Vision (ICCV)}, 2021.

\bibitem{fastsam}
Xu~Zhao, Wenchao Ding, Yongqi An, Yinglong Du, Tao Yu, Min Li, Ming Tang, and Jinqiao Wang.
\newblock Fast segment anything.
\newblock {\em arXiv preprint arXiv:2306.12156}, 2023.

\bibitem{zhou2023edgesam}
Chong Zhou, Xiangtai Li, Chen~Change Loy, and Bo~Dai.
\newblock Edgesam: Prompt-in-the-loop distillation for on-device deployment of sam.
\newblock {\em arXiv preprint arXiv:2312.06660}, 2023.

\bibitem{zhou2023bcinet}
Wujie Zhou, Yuchun Yue, Meixin Fang, Xiaohong Qian, Rongwang Yang, and Lu~Yu.
\newblock Bcinet: Bilateral cross-modal interaction network for indoor scene understanding in rgb-d images.
\newblock {\em Information Fusion}, 94:32--42, 2023.

\end{thebibliography}


\appendix
\section{Appendix}

\subsection{Detailed Results on Instance/Panoptic/Semantic Segmentation}
In this section, we present here our detailed performance across various tasks and datasets.
Table \ref{table:detail_instance_scannet},
Table \ref{table:detail_instance_replica} and Table\ref{table:detail_instance_scannetpp} respectively show our detailed performance on the 3D instance segmentation task for ScanNet v2, ScanNet200, and ScanNet++.
Table \ref{table:detail_panoptic_scannet} and Table \ref{table:detail_panoptic_replica} present our detailed performance on ScanNet v2 and Replica, respectively,
in comparison with the open-vocabulary panoptic segmentation method PVLFF, which utilizes NVS.
Table \ref{table:detail_semantic_scannet} and Table \ref{table:detail_semantic_repica} primarily compare our method with the open-vocabulary semantic segmentation method based on 2D images on ScanNet v2 and Replica.
Table \ref{table:detail_semantic_scannet_3d} shows a detailed comparison with the open-vocabulary 3D semantic segmentation method based on 2D-3D merging.

\begin{table}[h]
\centering
\resizebox{\linewidth}{!}{
\begin{tabular}{l|ccccccc}
\toprule
\textbf{Scene Name} & \textbf{SAM3D}\cite{yang2023sam3d} & \textbf{SAI3D}\cite{yin2024sai3d} & \textbf{Mask Clust.}\cite{yan2024maskclustering} & \textbf{Open3DIS}\cite{nguyen2023open3dis} & \textbf{Gaussian Group.}\cite{gaussian_grouping} & \textbf{Cues3D} & \textbf{Cues3D+SP} \\
\midrule
scene0050\_02 & 7.5 & \textbf{22.9} & 19.9 & 26.3 & 1.0 & 19.8 & 28.9 \\ 
scene0144\_01 & 9.2 & 16.9 & 36.1 & \textbf{37.4} & 2.5 & 26.0 & 37.0 \\ 
scene0221\_01 & 6.7 & \textbf{45.7} & 31.5 & 22.9 & 1.2 & 11.7 & 20.3 \\ 
scene0300\_01 & 8.8 & 33.0 & 24.1 & 23.5 & 0.9 & 25.2 & \textbf{35.1} \\ 
scene0354\_00 & 1.8 & 33.9 & \textbf{43.3} & 31.8 & 0.0 & 17.5 & 32.8 \\ 
scene0389\_00 & 17.2 & 29.6 & 47.3 & \textbf{54.4} & 8.1 & 44.9 & 60.0 \\ 
scene0423\_02 & 25.9 & \textbf{100.0} & \textbf{100.0} & \textbf{100.0} & 21.8 & \textbf{100.0} & \textbf{100.0} \\ 
scene0427\_00 & 54.3 & 36.5 & 53.0 & \textbf{87.2} & 12.0 & 52.8 & 70.4 \\ 
scene0494\_00 & 5.0 & 44.7 & 46.2 & 45.2 & 1.3 & 37.0 & \textbf{54.2} \\ 
scene0616\_00 & 15.1 & 22.4 & 47.4 & 29.2 & 1.3 & 42.0 & \textbf{52.4} \\ 
scene0645\_02 & 2.9 & 32.1 & 20.2 & 28.3 & 0.7 & 19.5 & \textbf{33.0} \\ 
scene0693\_00 & 17.2 & 0.8 & 22.1 & 27.0 & 1.5 & 19.6 & \textbf{25.0} \\ \midrule
\textbf{Average} & 14.3 & 34.9 & 40.9 & 42.8 & 4.4 & 34.7 & \textbf{45.8} \\ \bottomrule
\end{tabular}}
\caption{\textbf{Category-agnostic 3D instance segmentation} on ScanNet v2.
We report the mean of AP from 50\% to 95\% at 5\% intervals.
Best results are in bold.}
\label{table:detail_instance_scannet}
\end{table}

\begin{table}[h]
\centering
\resizebox{\linewidth}{!}{
\begin{tabular}{l|ccccccc}
\toprule
\textbf{Scene Name} & \textbf{SAM3D}\cite{yang2023sam3d} & \textbf{SAI3D}\cite{yin2024sai3d} & \textbf{Mask Clust.}\cite{yan2024maskclustering} & \textbf{Open3DIS}\cite{nguyen2023open3dis} & \textbf{Gaussian Group.}\cite{gaussian_grouping} & \textbf{Cues3D} & \textbf{Cues3D+SP} \\  
\midrule
scene0050\_02 & 8.4 & 29.1 & 27.8 & \textbf{38.9} & 1.2 & 25.6 & 38.4 \\ 
scene0144\_01 & 7.1 & 15.8 & 28.7 & 27.3 & 1.8 & 22.3 & \textbf{31.1} \\ 
scene0221\_01 & 2.9 & \textbf{35.1} & 25.8 & 26.0 & 1.0 & 15.7 & 31.6 \\ 
scene0300\_01 & 12.3 & 38.2 & 30.2 & 27.5 & 0.9 & 29.9 & \textbf{44.1} \\ 
scene0354\_00 & 4.4 & 33.4 & \textbf{45.8} & 34.5 & 1.6 & 22.3 & 38.3 \\ 
scene0389\_00 & 15.7 & 22.4 & 51.0 & 52.1 & 5.4 & 44.8 & \textbf{64.8} \\ 
scene0423\_02 & 25.9 & \textbf{100.0} & \textbf{100.0} & \textbf{100.0} & 12.9 & \textbf{100.0} & \textbf{100.0} \\ 
scene0427\_00 & 45.9 & 33.1 & 46.4 & \textbf{64.8} & 6.7 & 46.9 & 60.9 \\ 
scene0494\_00 & 8.0 & 47.0 & 49.4 & 40.3 & 2.9 & 39.7 & \textbf{53.3} \\ 
scene0616\_00 & 12.7 & 18.6 & 45.4 & 28.7 & 1.4 & 33.7 & \textbf{46.6} \\ 
scene0645\_02 & 2.8 & 21.7 & 26.4 & 28.6 & 0.7 & 22.6 & \textbf{40.6} \\ 
scene0693\_00 & 7.2 & 16.3 & 16.7 & 24.3 & 2.6 & 15.2 & \textbf{25.5} \\ \midrule
\textbf{Average} & 12.8 & 34.2 & 41.1 & 41.1 & 3.3 & 34.9 & \textbf{47.9} \\ \bottomrule
\end{tabular}}
\caption{\textbf{Category-agnostic 3D instance segmentation} on ScanNet200. We report the mean of AP from 50\% to 95\% at 5\% intervals.
Best results are in bold.}
\label{table:detail_instance_replica}
\end{table}

\begin{table}[h]
\centering
\resizebox{\linewidth}{!}{
\begin{tabular}{l|cccccc}
\toprule
\textbf{Scene Name} & \textbf{SAM3D} \cite{yang2023sam3d} & \textbf{SAI3D} \cite{yin2024sai3d} & \textbf{Mask Clust.} \cite{yan2024maskclustering} & \textbf{Open3DIS} \cite{nguyen2023open3dis} & \textbf{Cues3D} & \textbf{Cues3D+SP} \\
\midrule
7b6477cb95 & 5.0 & 14.1 & \textbf{23.9} & 16.0 & 12.7 & 17.5 \\ 
825d228aec & 34.0 & 15.2 & 28.6 & 29.4 & 24.5 & \textbf{29.7} \\ 
a24f64f7fb & 15.8 & 19.0 & 41.6 & 26.6 & 42.1 & \textbf{48.6} \\ 
1ada7a0617 & 6.8 & 19.0 & \textbf{40.3} & 27.7 & 20.7 & 28.0 \\ 
5748ce6f01 & 12.1 & 15.9 & \textbf{41.4} & 27.9 & 35.8 & 37.6 \\ 
40aec5fffa & 3.0 & 14.0 & \textbf{24.6} & 18.2 & 20.9 & 23.2 \\ 
bcd2436daf & 8.6 & 22.4 & 42.7 & 23.7 & 41.9 & \textbf{50.0} \\ 
6115eddb86 & 7.6 & 16.2 & 36.4 & 27.2 & 25.8 & \textbf{40.7} \\ \midrule
\textbf{Average} & 11.6 & 17.0 & \textbf{34.9} & 24.6 & 28.1 & 34.4 \\ \bottomrule
\end{tabular}}
\caption{\textbf{Category-agnostic 3D instance segmentation} on ScanNet++.
We report the mean of AP from 50\% to 95\% at 5\% intervals.
Best results are in bold.}
\label{table:detail_instance_scannetpp}
\end{table}

\begin{table}[h]
\scriptsize
\centering
\begin{tabular}{l|cccccc}
\toprule
\multirow{2.5}{*}{\textbf{Scene Name}} & \multicolumn{2}{c}{\textbf{PVLFF(LSeg)}\cite{pvlff}} & \multicolumn{2}{c}{\textbf{Cues3D(CLIP)}} & \multicolumn{2}{c}{\textbf{Cues3D(FC-CLIP)}} \\
\cmidrule(r){2-3} \cmidrule(r){4-5} \cmidrule(r){6-7}
 & $\mathrm{\textbf{PQ}}^{\text{\textbf{scene}}}$ & \textbf{mIoU} & $\mathrm{\textbf{PQ}}^{\text{\textbf{scene}}}$ & \textbf{mIoU} & $\mathrm{\textbf{PQ}}^{\text{\textbf{scene}}}$ & \textbf{mIoU} \\
\midrule
scene0050\_02 & 7.0 & 24.5 & \textbf{39.2} & 45.9 & 36.4 & \textbf{56.0} \\
scene0144\_01 & 10.0 & 40.6 & \textbf{35.4} & 40.8 & 31.2 & \textbf{52.6} \\
scene0221\_01 & 6.4 & 20.9 & 11.1 & 15.1 & \textbf{12.8} & \textbf{23.8} \\
scene0300\_01 & 15.9 & 50.0 & 32.2 & 45.0 & \textbf{41.8} & \textbf{66.2} \\
scene0354\_00 & 20.1 & 46.9 & 11.4 & 22.8 & \textbf{35.5} & \textbf{53.8} \\
scene0389\_00 & 15.2 & 41.7 & 44.8 & 50.4 & \textbf{51.4} & \textbf{67.3} \\
scene0423\_02 & 22.6 & 60.7 & 44.2 & 59.9 & \textbf{64.4} & \textbf{92.2} \\
scene0427\_00 & 20.0 & 50.3 & 47.2 & 48.2 & \textbf{59.1} & \textbf{78.3} \\
scene0494\_00 & 17.4 & 53.5 & 34.7 & 43.3 & \textbf{43.6} & \textbf{62.7} \\
scene0616\_00 & 9.7 & 36.2 & \textbf{31.7} & 41.3 & 31.6 & \textbf{52.9} \\
scene0645\_02 & 7.3 & 25.5 & \textbf{20.0} & 29.6 & 19.8 & \textbf{38.8} \\
scene0693\_00 & 6.6 & 25.5 & 14.1 & 19.3 & \textbf{18.8} & \textbf{35.5} \\
\midrule
\textbf{Average} & 13.2 & 39.7 & 30.5 & 38.5 & \textbf{37.2} & \textbf{56.7} \\
\bottomrule
\end{tabular}
\caption{\textbf{Open-vocabuary 3D panoptic segmentation} on ScanNet v2.
Best results are in bold.}
\label{table:detail_panoptic_scannet}
\end{table}

\begin{table}[h]
\scriptsize
\centering
\begin{tabular}{l|cccccc}
\toprule
\multirow{2.5}{*}{\textbf{Scene Name}} & \multicolumn{2}{c}{\textbf{PVLFF(LSeg)}\cite{pvlff}} & \multicolumn{2}{c}{\textbf{Cues3D(CLIP)}} & \multicolumn{2}{c}{\textbf{Cues3D(FC-CLIP)}} \\
\cmidrule(r){2-3} \cmidrule(r){4-5} \cmidrule(r){6-7}
 & $\mathrm{\textbf{PQ}}^{\text{\textbf{scene}}}$ & \textbf{mIoU} & $\mathrm{\textbf{PQ}}^{\text{\textbf{scene}}}$ & \textbf{mIoU} & $\mathrm{\textbf{PQ}}^{\text{\textbf{scene}}}$ & \textbf{mIoU} \\
\midrule
office\_0 & 7.0 & 18.9 & 10.5 & 25.3 & \textbf{15.1} & \textbf{35.5} \\
office\_2 & 12.4 & 29.9 & 15.7 & 27.1 & \textbf{21.8} & \textbf{33.2} \\
office\_4 & 13.9 & 24.0 & 18.3 & 36.4 & \textbf{22.3} & \textbf{36.9} \\
room\_0   & 8.1 & 27.2 & 11.0 & 28.2 & \textbf{15.5} & \textbf{46.0} \\
room\_1   & 11.0 & 27.5 & 10.6 & 22.8 & \textbf{11.4} & \textbf{50.1}\\
room\_2   & 13.3 & 29.6 & 10.2 & 24.3 & \textbf{11.3} & \textbf{49.7} \\
\midrule
\textbf{Average} & 11.0 & 26.2 & 12.7 & 27.3 & \textbf{16.2} & \textbf{41.9} \\
\bottomrule
\end{tabular}
\caption{\textbf{Open-vocabuary 3D panoptic segmentation} method PVLFF on Replica.}
\label{table:detail_panoptic_replica}
\end{table}

\begin{table}[h]
  \centering
  \resizebox{\linewidth}{!}{%
  \begin{tabular}{l|cccccccccc}
    \toprule
    \multirow{2.5}{*}{\textbf{Scene Name}} & \multicolumn{2}{c}{\textbf{LERF(CLIP)}\cite{kerr2023lerf}} & \multicolumn{2}{c}{\textbf{LangSplat(CLIP)}\cite{qin2023langsplat}} & \multicolumn{2}{c}{\textbf{PVLFF(LSeg)}\cite{pvlff}} & \multicolumn{2}{c}{\textbf{Cues3D(CLIP)}} & \multicolumn{2}{c}{\textbf{Cues3D(FC-CLIP)}} \\
    \cmidrule(r){2-3} \cmidrule(r){4-5} \cmidrule(r){6-7} \cmidrule(r){8-9} \cmidrule(r){10-11}
    & \textbf{mIoU} & \textbf{mAcc} & \textbf{mIoU} & \textbf{mAcc} & \textbf{mIoU} & \textbf{mAcc} & \textbf{mIoU} & \textbf{mAcc} & \textbf{mIoU} & \textbf{mAcc} \\
    \midrule
    scene0050\_02 & 22.9 & 48.2 & 17.4 & 29.5 & 24.5 & 31.6 & 45.9 & 62.8 & \textbf{56.0} & \textbf{71.5} \\
    scene0144\_01 & 26.8 & 40.3 & 16.4 & 28.0 & 40.6 & 58.4 & 40.8 & 55.3 & \textbf{52.6} & \textbf{63.6} \\
    scene0221\_01 & 7.4  & 20.1 & 7.4  & 17.7 & 20.9 & 27.8 & 15.1 & 27.3 & \textbf{23.8} & \textbf{36.6} \\
    scene0300\_01 & 37.8 & 59.8 & 28.3 & 46.6 & 50.0 & 56.9 & 45.0 & 57.2 & \textbf{66.2} & \textbf{72.4} \\
    scene0354\_00 & 23.6 & 55.5 & 11.6 & 24.4 & 46.9 & 54.2 & 22.8 & 33.2 & \textbf{53.8} & \textbf{61.7} \\
    scene0389\_00 & 31.6 & 53.0 & 18.1 & 31.3 & 41.7 & 50.3 & 50.4 & 70.3 & \textbf{67.3} & \textbf{79.9} \\
    scene0423\_02 & 30.4 & 45.6 & 55.9 & 65.4 & 60.7 & 65.4 & 59.9 & 65.1 & \textbf{92.2} & \textbf{79.9} \\
    scene0427\_00 & 34.7 & 59.5 & 38.2 & 55.2 & 50.3 & 60.0 & 48.2 & 61.1 & \textbf{78.3} & \textbf{84.7} \\
    scene0494\_00 & 37.8 & 55.4 & 18.0 & 34.7 & 53.5 & 61.0 & 43.3 & 55.8 & \textbf{62.7} & \textbf{79.0} \\
    scene0616\_00 & 29.4 & 57.4 & 13.3 & 24.7 & 36.2 & 46.0 & 41.3 & 55.7 & \textbf{52.9} & \textbf{67.3} \\
    scene0645\_02 & 10.8 & 27.0 & 4.8  & 10.7 & 25.5 & 34.0 & 29.7 & 43.8 & \textbf{38.8} & \textbf{53.7} \\
    scene0693\_00 & 13.5 & 31.9 & 16.7 & 36.7 & 25.5 & 36.0 & 19.3 & 38.5 & \textbf{35.5} & \textbf{49.0} \\
    \midrule
    \textbf{Average} & 25.6 & 46.1 & 20.5 & 33.7 & 39.7 & 48.5 & 38.5 & 52.2 & \textbf{56.7} & \textbf{66.6} \\
    \bottomrule
  \end{tabular}}
  \caption{\textbf{Open-vocabulary 3D semantic segmentation} based on 2D images on ScanNet v2.}
  \label{table:detail_semantic_scannet}
\end{table}

\begin{table}[h]
  \centering
  \resizebox{\linewidth}{!}{
    \begin{tabular}{l|cccccccccc}
      \toprule
      \multirow{2.5}{*}{\textbf{Scene Name}} & \multicolumn{2}{c}{\textbf{LERF(CLIP)}\cite{kerr2023lerf}} & \multicolumn{2}{c}{\textbf{LangSplat(CLIP)}\cite{qin2023langsplat}} & \multicolumn{2}{c}{\textbf{PVLFF(LSeg)}\cite{pvlff}} & \multicolumn{2}{c}{\textbf{Cues3D(CLIP)}} & \multicolumn{2}{c}{\textbf{Cues3D(FC-CLIP)}} \\
      \cmidrule(r){2-3} \cmidrule(r){4-5} \cmidrule(r){6-7} \cmidrule(r){8-9} \cmidrule(r){10-11}
      & \textbf{mIoU} & \textbf{mAcc} & \textbf{mIoU} & \textbf{mAcc} & \textbf{mIoU} & \textbf{mAcc} & \textbf{mIoU} & \textbf{mAcc} & \textbf{mIoU} & \textbf{mAcc} \\
      \midrule
      office\_0 & 15.5 & 27.5 & 17.8 & 23.7 & 18.9 & 26.7 & 25.3 & 28.4 & \textbf{35.5} & \textbf{48.9} \\
      office\_2 & 15.5 & 34.0 & 19.5 & 33.2 & 29.9 & 34.3 & 27.1 & 37.6 & \textbf{33.2} & \textbf{47.8} \\
      office\_4 & 21.2 & 44.7 & 8.0  & 19.8 & 24.0 & 32.1 & 36.4 & 41.9 & \textbf{36.9} & \textbf{48.8} \\
      room\_0   & 20.1 & 35.4 & 16.3 & 24.6 & 27.2 & 36.7 & 28.2 & 32.4 & \textbf{46.0} & \textbf{53.2} \\
      room\_1   & 15.8 & 37.9 & 15.3 & 22.9 & 27.5 & 39.9 & 22.8 & 34.1 & \textbf{50.1} & \textbf{55.6} \\
      room\_2   & 14.4 & 30.4 & 7.9  & 15.1 & 29.6 & 40.3 & 24.3 & 38.4 & \textbf{49.7} & \textbf{64.2} \\
      \midrule
      \textbf{Average} & 17.1 & 35.0 & 14.1 & 23.2 & 26.2 & 35.0 & 27.3 & 35.5 & \textbf{41.9} & \textbf{53.1} \\
      \bottomrule
    \end{tabular}
  }
  \caption{\textbf{Open-vocabulary 3D semantic segmentation} method based on 2D images on Replica.}
  \label{table:detail_semantic_repica}
\end{table}

\begin{table}[h]
  \centering
  \resizebox{\linewidth}{!}{%
  \begin{tabular}{l|cccccccc}
    \toprule
    \multirow{2.5}{*}{\textbf{Scene Name}} & \multicolumn{2}{c}{\textbf{OpenScene(Ensemble)}\cite{peng2023openscene}} & \multicolumn{2}{c}{\textbf{Cues3D(CLIP)}} & \multicolumn{2}{c}{\textbf{Cues3D(OpenSeg)}} & \multicolumn{2}{c}{\textbf{Cues3D(FC-CLIP)}} \\
    
    \cmidrule(r){2-3} \cmidrule(r){4-5} \cmidrule(r){6-7} \cmidrule(r){8-9}
    & \textbf{mIoU} & \textbf{mAcc} & \textbf{mIoU} & \textbf{mAcc} & \textbf{mIoU} & \textbf{mAcc} & \textbf{mIoU} & \textbf{mAcc} \\
    \midrule
    scene0050\_02 & 26.3 & 39.3 & 17.6 & 29.6 & 22.2 & 31.7 & \textbf{27.1} & \textbf{56.1} \\
    scene0144\_01 & 31.6 & 45.1 & 33.6 & 43.6 & 34.4 & 57.4 & \textbf{40.8} & \textbf{74.8} \\
    scene0221\_01 & \textbf{32.9} & 51.3 & 18.2 & 37.7 & 27.7 & 47.5 & 30.0 & \textbf{54.0} \\
    scene0300\_01 & 28.6 & 45.3 & 24.4 & 44.7 & 37.8 & 62.5 & \textbf{40.0} & \textbf{66.4} \\
    scene0354\_00 & 26.6 & 50.6 & 5.7  & 9.4  & \textbf{35.2} & 54.5 & 31.3 & \textbf{62.7} \\
    scene0389\_00 & 39.5 & 63.8 & 48.6 & 69.1 & 47.1 & 57.8 & \textbf{63.4} & \textbf{80.5} \\
    scene0423\_02 & 43.5 & 58.7 & 21.1 & 38.6 & 42.8 & 73.3 & \textbf{56.9} & \textbf{96.7} \\
    scene0427\_00 & 33.6 & 60.8 & 19.1 & 36.5 & 38.2 & 58.5 & \textbf{52.6} & \textbf{71.6} \\
    scene0494\_00 & 31.7 & 50.1 & 17.5 & 42.6 & 33.5 & 53.9 & \textbf{45.9} & \textbf{64.7} \\
    scene0616\_00 & 32.3 & 53.7 & 31.6 & 45.5 & 35.6 & 44.6 & \textbf{44.7} & \textbf{64.4} \\
    scene0645\_02 & 35.6 & 64.2 & 36.3 & 49.5 & 41.1 & 49.7 & \textbf{61.1} & \textbf{69.3} \\
    scene0693\_00 & \textbf{46.2} & \textbf{71.8} & 15.2 & 36.2 & 39.6 & 62.6 & 31.1 & 63.7 \\
    \midrule
    \textbf{Average} & 34.0 & 54.5 & 24.1 & 40.2 & 36.3 & 54.5 & \textbf{43.7} & \textbf{68.7} \\
    \bottomrule
  \end{tabular}}
  \caption{\textbf{Open-vocabulary 3D semantic segmentation} method based on 2D-3D merging on ScanNet v2.}
  \label{table:detail_semantic_scannet_3d}
\end{table}

\subsection{Panoptic Comparisons in the setting of PVLFF}
\label{appendix:compare_pvlff}
In the experimental setup of PVLFF \cite{pvlff},
only a subset of categories from the datasets is selected for evaluation.
To thoroughly assess the effectiveness of all methods,
our experimental setup includes all categories provided by the datasets.
For a more comprehensive comparison with PVLFF,
we perform evaluations using the same experimental settings as PVLFF.
Note that since PVLFF does not provide the configuration file for the ScanNet dataset,
we can only perform the experiment on the Replica dataset.\\

Table \ref{table:compare_pvlff} shows the panoptic results of Cues3D and PVLFF on the Replica dataset.
The conclusion is the same as our experiment in Sec. \ref{exp:sota}.
Cues3D outperforms PVLFF by 4.8\% on ${\text{PQ}}^{\text{scene}}$, 9.8\% on mIoU, and 6\% on mAcc.
The reason is that we use full supervision in the third phase instead of a clustering algorithm, resulting in less noise compared to PVLFF.
Additionally, when performing semantic projection,
we incoporate multi-view 2D semantic to an object, which avoid errors caused by single view prediction.

\subsection{Impact of Voxel Size}
\label{appendix:voxel_size}
\xf{
In order to explore the impact of voxel size on performance, we also conducted corresponding experiments to prove that the voxel size we adopted is reasonable.
As shown in Table \ref{table:voxel_size},
setting the voxel size to 0.05 yields the best performance.
Increasing the voxel size for higher efficiency results in a significant drop in AP25 and AP50 and cannot greatly speed up disambiguation.
}

\begin{table}[t]
\scriptsize
    \centering
    \caption{\xf{\textbf{Ablation on voxel size.}
    We present the performance on ScanNet200.}}
    \label{table:voxel_size}
    \begin{tabular}{lcccc}
        \toprule
        \textbf{Voxel Size} & \textbf{AP} & \textbf{AP25} & \textbf{AP50} & \textbf{Time}\\
        \midrule
        0.05  & \underline{34.9} & \textbf{61.2} & \textbf{76.9} & 2.7 minutes\\
        0.075 & \textbf{35.0} & \underline{56.2} & \underline{73.8} & 2.4 minutes\\
        0.1   & 23.5 & 40.4 & 64.7 & 2.2 minutes\\
        \bottomrule
    \end{tabular}
\end{table}

\subsection{Code and Data}
We provide code and partial data at this \href{https://drive.google.com/drive/folders/16NH9nDjr-b9qja1gk31XbxavT-SGCDEP?usp=sharing}{link}.
For the ``cues3d.zip'' file,
in the ``data'' folder,
we offer example data for one scene to demonstrate data organization.
Organizing data from other scenes in this structure should also enable the program to run properly.
The ``cues3d'' folder contains the training part of the network.
The ``instance\_disambiguation'' folder holds the algorithm described for Instance Disambiguation.
The ``semantic\_seg'' folder contains relevant programs for feature extraction and multi-view semantic voting.
In ``cues3d/outputs'', we provide the running results of CropFormer for each scene.
For details on environment setup and running the program,
please refer to the ``README.md''. 
The ``vis\_result.zip'' file provides visualization results.


\begin{table}[h]
\scriptsize
\centering
\caption{\textbf{Open-vocabulary 3D panoptic segmentation results using the same setting as PVLFF.}
We present the performance of on Replica.
The denoised semantic segmentation results of PVLFF are reported in parentheses.}
\label{table:compare_pvlff}
\begin{tabular}{l ccccc}
\toprule
\textbf{Method}   & \textbf{Image Feature}   & \textbf{PQ}$^{\textbf{scene}}$ & \textbf{mIoU} & \textbf{mAcc} \\ 
\midrule
PVLFF \cite{pvlff}  &  LSeg   & 48.1 & 54.1 (56.7) & 64.1 (65.4) \\ 
\textbf{Cues3D}     &  LSeg  & \textbf{52.9} & \textbf{63.9} & \textbf{70.1} \\ 
\bottomrule
\end{tabular}
\end{table}

\begin{figure}[h]
\centering
\includegraphics[width=0.57\textwidth]{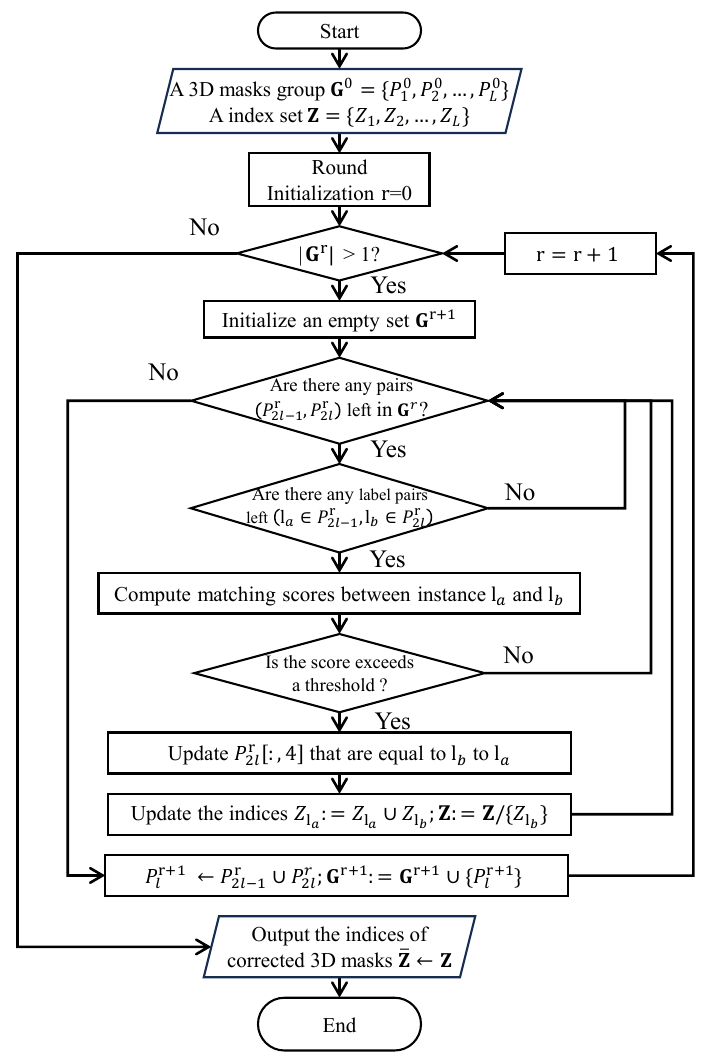}
\caption{\xf{\textbf{Flow diagram of Instance Disambiguation.}}}
\label{fig:flow}
\end{figure}

\xf{
\subsection{Flow diagram of Instance Disambiguation}
\label{sec:flow}
Fig. \ref{fig:flow} shows the detailed process of the instance disambiguation part more clearly.
We simplify the calculation process of the matching score to ensure that it is easier to understand.}

\end{document}